\pdfoutput=1

\documentclass[11pt]{article}

\usepackage {acl}

\usepackage{times}
\usepackage{latexsym}
\usepackage{booktabs}
\usepackage{amssymb}
\usepackage{amsmath}
\usepackage{bbm}
\usepackage{mathtools}
\usepackage{url}
\usepackage{float}
\usepackage{multirow}
\usepackage{subcaption}
\usepackage{xcolor}
\usepackage{xspace}
\usepackage{ulem} 
\usepackage{soul}
\usepackage{svg}

\usepackage[T1]{fontenc}

\usepackage[utf8]{inputenc}

\usepackage{microtype}

\usepackage{inconsolata}

\newcommand{\model}{\textsc{MIDGARD}\xspace}

\title{MIDGARD: Self-Consistency Using Minimum Description Length \\for Structured Commonsense Reasoning}

\author{Inderjeet Nair \and Lu Wang \\
  University of Michigan, Ann Arbor, MI \\
  \texttt{\{inair, wangluxy\}@umich.edu} \\} 

\begin{document}
\maketitle
\begin{abstract}
We study the task of conducting structured reasoning as generating a reasoning graph from natural language input using large language models (LLMs). 
Previous approaches have explored various prompting schemes, yet they suffer from error propagation due to the autoregressive nature and single-pass-based decoding, which lack error correction capability. Additionally, relying solely on a single sample may result in the omission of true nodes and edges.
To counter this, we draw inspiration from \textit{self-consistency} (SC), which involves sampling a diverse set of reasoning chains and taking the majority vote as the final answer. 
%
To tackle the substantial challenge of applying SC on generated graphs, we propose \textbf{\model} (\ul{MI}nimum \ul{D}escription length \ul{G}uided \ul{A}ggregation of \ul{R}easoning in \ul{D}irected acyclic graph) that leverages Minimum Description Length (MDL)-based formulation to identify consistent properties among the different graph samples generated by an LLM.
This formulation helps reject properties that appear in only a few samples, which are likely to be erroneous, while enabling the inclusion of missing elements without compromising precision. Our method demonstrates superior performance than comparisons across various structured reasoning tasks, including argument structure extraction, explanation graph generation, inferring dependency relations among actions for everyday tasks, and semantic graph generation from natural texts.
\end{abstract}

\begin{figure*}
    \centering
    \includegraphics[scale=0.63]{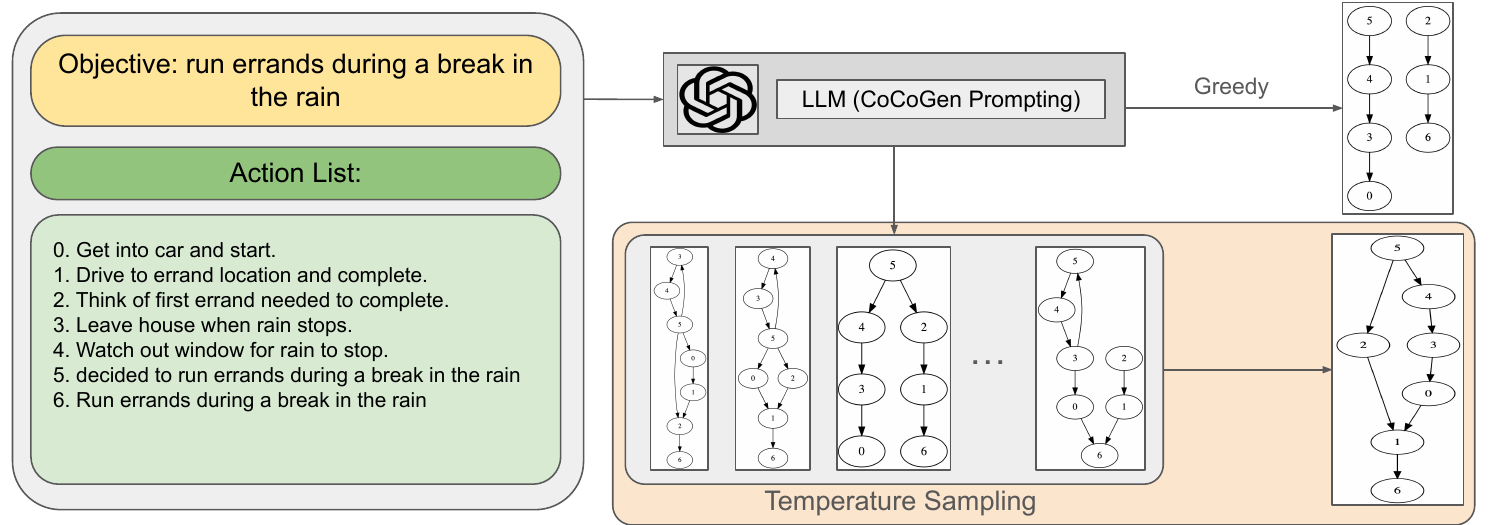}
    \caption{
    Comparison of \model with \textsc{CoCoGen}. In this example, our objective is to infer dependency relations among items in the "Action List" to achieve the specified "Objective". \textsc{CoCoGen} uses greedy decoding and exhibits errors in the output, e.g., "decided to run errands during a break in the rain" is not connected with "Drive to errand location and complete". 
    In contrast, our approach \model (within the \colorbox[HTML]{FCE5CD}{orange} rectangle) aggregates relevant information across different samples, resulting in more accurate inference. 
    For this example, our algorithm improved the performance of greedy decoding from $\mathbf{66.7}$ to $\mathbf{85.7}$ in edge $F_1$-score.
    }
    \label{fig:cocogen_comparison}
\end{figure*}

\section{Introduction}

While large language models (LLMs) have showcased impressive performance in few-shot/zero-shot scenarios across diverse reasoning tasks~\cite{brown2020language,chen2021evaluating,rae2022scaling,hoffmann2022training,chowdhery2022palm}, it is still challenging to apply these models for structured commonsense reasoning which involves generating task-specific reasoning as a graph, such as extracting argument structures
from argumentative text~\cite{stab2017parsing,hua-etal-2019-argument,mayer2020transformer,hua-wang-2022-efficient,qiao2022reasoning}, generating structured explanations that lay out commonsense knowledge to connect an argument to a belief~\cite{saha-etal-2021-explagraphs}, and inferring dependencies among events for everyday activities~\cite{sakaguchi-etal-2021-proscript-partially}. 

There are two main challenges for \textbf{structured reasoning} tasks. 
(1) \textit{Style discrepancy}: Conventional approaches for structured response generation represent the graphs as flattened strings~\cite{madaan-yang-2021-neural,madaan-etal-2021-give,sakaguchi-etal-2021-proscript-partially,saha-etal-2021-explagraphs}, leading to subpar performance due to output style mismatch~\cite{madaan-etal-2022-language}.  
(2) \textit{Error propagation}: Any incorrect decisions made earlier in the autoregressive decoding process can influence later generation steps~\cite{yao2023tree}. 
Recently, \citet{madaan-etal-2022-language} propose \textsc{CoCoGen} to address the issue of style mismatch in structured reasoning tasks, by using programming scripts as prompts for LLMs. 
It still suffers from error propagation, since it generates variable declarations and function calls in order to describe the nodes and edges within the graph. Any error in these declarations/calls can affect the subsequent generations.



To address these issues, we take inspiration from the \textit{self-consistency} (SC)~\cite{wang2023selfconsistency} strategy that samples diverse reasoning paths and then takes a majority vote as the final answer. 
The intuition behind SC is that sampling distinct reasoning chains leads to higher confidence in the correctness of a consistent answer. 
Therefore, we hypothesize that sampling diverse graphs from an LLM can construct a more accurate aggregate graph and alleviate error propagation for structured reasoning tasks as any errors made in one sample are less likely to persist across all the generated graphs. 

A crucial distinction between SC and our desideratum is that SC focuses exclusively on commonsense reasoning tasks~\cite{ling-etal-2017-program,clark2018think,cobbe2021training,patel-etal-2021-nlp,geva2021did} with \textbf{scalar answer spaces}. 
In contrast, we aim to \textbf{merge multiple graphs}, each representing a collection of unordered sets (nodes and edges). It is unclear how to apply majority vote to aggregate distinct sets of nodes and edges. In particular, it would be critical to filter out inaccurate nodes and edges in our setup.

To achieve this, we propose \textbf{\model}\footnote{Our code is publicly available at \url{https://github.com/launchnlp/MIDGARD}.}, based on \ul{MI}nimum \ul{D}escription length \ul{G}uided \ul{A}ggregation of \ul{R}easoning in \ul{D}irected acyclic graph. 
We employ the principle of \textit{minimum description length} (MDL)~\cite{RISSANEN1978465} which seeks to find the hypothesis with shortest description length of the observations. While MDL has been implemented for model selection~\cite{grunwald2005minimum}, causal structure learning~\cite{lam1993using,lam1994using}, data clustering~\cite{rissanen2000mdl}, and dimensionality reduction~\cite{bruni2022short}, to the best of our knowledge, its use in automatically merging graph samples has never been explored before. 
Assuming that graph properties consistent across multiple generated samples are more likely to be accurate, we define the description length of a graph sample as the weighted sum of the transformations required to convert a hypothesis into the given sample. By constructing a hypothesis that minimizes the description length across all the generated samples, our solution encourages the inclusion of graph properties that were present in many samples, while rejecting properties that were only present in a few samples which are likely to be erroneous. Figure~\ref{fig:cocogen_comparison} shows an example of how our approach reduces errors compared to relying solely on a single greedy generation. 
Empirical results on four different structured reasoning tasks, including argument structure extraction, structured explanation construction, and goal-oriented script generation and semantic graph generation, on eight benchmarks show that \model can outperform competitive baseline and model variants, demonstrating its strong generalizability.

\section{Background and Notations}
\label{sec:objective_formulation}

In structured reasoning, a labeled data point is denoted as $(\mathcal{T}, \mathcal{G})$, where $\mathcal{T}$ represents the input and $\mathcal{G}$ is the task-specific graph output that captures the necessary reasoning knowledge. 
For example, in the task of argumentative structure extraction~\cite{stab2017parsing}, $\mathcal{T}$ can be an essay, and $\mathcal{G}$ represents the associated argumentative structure. 

To solve this task, we employ LLM in the \textit{few-shot prompting mode} where $N$ labelled data-points $\{\mathcal{T}_i, \mathcal{G}_i\}_{i=1}^N$
are fed as in-context prompt to the model to infer the output for a test input $\mathcal{T}$. In accordance with the \textsc{CoCoGen} approach, we construct the in-context prompt as follows:
$p = \mathcal{T}_1 \,\oplus\, \mathcal{G}_1^c \,\cdot\, \mathcal{T}_2\, \oplus \,\mathcal{G}_2^c \,\cdot\, \dots\, \cdot\, \mathcal{T}_N\, \oplus\, \mathcal{G}_N^c$ where $\mathcal{G}_i^c$ is a semantically equivalent representation of $\mathcal{G}_i$ written in a generally purpose programming language like Python and $\cdot$ ($\oplus$) represents inter (intra)-instance separator. $\mathbb{P}_c\left(\cdot, \mathcal{T}\right)$ represents the generative distribution of the LLM for the prompt $p$ and $\mathcal{T}$. While \textsc{CoCoGen} relies on a single generation obtained from $\mathbb{P}_c\left(\cdot, \mathcal{T}\right)$, our approach utilizes this to generate multiple graph samples $\{\mathcal{G'}_i\}_{i=1}^T \sim \mathbb{P}_c\left(\cdot, \mathcal{T}\right)$ and then aggregates them into a single output $\mathcal{G}$. Our novelty lies in the development of a novel and generic aggregation algorithm for the task of reasoning graph generation. This algorithm leads to significantly improved performance across multiple tasks.

\section{The \model Method}
\label{sec:mdl_for_objective_formulation}

\model is based on the principle of minimum description length (MDL), which succinctly captures the regularities in the given data by finding the hypothesis with the shortest description length. 
In graph aggregation, MDL can be used as a self-consistency strategy to merge multiple reasoning graph samples into a single aggregate graph. 
The \textit{core idea} is to define a description length for each graph sample, which is proportional to the number of transformations required to convert a hypothesis into the given sample. By minimizing the description length for samples $\{\mathcal{G}'_i\}_{i=1}^T$, MDL encourages the inclusion of graph properties that are common across different samples. 
This means that properties that appear frequently in the generated samples are more likely to be accurate and reflect the underlying structure. Conversely, properties that are only present in a few samples tend to be wrong. 

In many structured reasoning tasks, the graphs typically do not have singleton nodes. For example, in argumentative structures of essays~\cite{stab2017parsing}, nodes are either supported or attacked by other nodes, or they themselves support or attack other nodes. 
We begin by defining the description length of a graph $\mathcal{G}'$ based on the hypothesis $\mathcal{G}$ when the graphs do not contain singleton nodes in \S \ref{sec:defining_description_length}. Next, we derive the expression for the expected description length in \S\ref{sec:eqn_derivation} assuming that $\mathcal{G}'$ is sampled from an LLM. Based on this, we formulate an objective that aims to minimize the expected description length of the sampled graphs $\{\mathcal{G'}_i\}_{i=1}^T \sim \mathbb{P}_c\left(\cdot, \mathcal{T}\right)$ in \S\ref{sec:hypothesis_selection}. We conclude this section by proposing modifications to the objective to address the scenario where the graph can have singleton nodes in \S\ref{sec:generic_graph_objective_formulation}.

\subsection{Defining Description Length}
\label{sec:defining_description_length}
We define \textbf{description length} of $\mathcal{G}'$ w.r.t. hypothesis graph $\mathcal{G}$ as follows:
\begin{equation}
    \label{eqn:dl_no_singleton}
    \Delta_\mathbf{E}(\mathcal{G}', \mathcal{G}, \lambda) = \lambda \cdot a + (1 - \lambda) \cdot d
\end{equation}
where $a$ represents the number of new edges to be added to $\mathcal{G}$, and $d$ is the number of edges to be deleted from $\mathcal{G}$ to convert it to $\mathcal{G}$'. We introduce the hyperparameter $\lambda$, which can be interpreted as the number of bits needed to describe a single addition, when $(1 - \lambda)$ bits are needed to describe a single deletion. The subscript $\mathbf{E}$ in Eq.~\ref{eqn:dl_no_singleton} indicates that only edge transformations are considered when calculating the description length. Since these graphs do not have isolated nodes and each node is associated with at least one edge, the description length of $\mathcal{G}'$ can be precisely captured using edge transformations alone.

The definition in Eq.~\ref{eqn:dl_no_singleton} is inspired by the formulation proposed by \citet{lam1994using}
, who applied it for refining causal graphs based on new data but not for the task of graph aggregation. While \citet{lam1994using} assigned equal bit requirements for describing a single addition and deletion, our empirical results demonstrate the significance of assuming different bit requirements to achieve enhanced performance. 

\subsection{Expected Description Length}
\label{sec:eqn_derivation}
We denote the set of nodes and edges associated with $\mathcal{G}$ as $N(\mathcal{G})$ and $E(\mathcal{G})$ respectively. Similarly, we define $\mathbf{N}$ and $\mathbf{E}$ as the sets of all possible nodes and edges, such that each edge (node) in $\mathcal{G}'$ and $\mathcal{G}$ belongs to $\mathbf{E}$ ($\mathbf{N}$). 
Taking the expectation of Eq.~\ref{eqn:dl_no_singleton} w.r.t. $\mathcal{G}' \sim \mathbb{P}_c(\dot, \mathcal{T})$

{\small
\begin{align}
    &\mathbb{E}_{\mathcal{G}'}\left[ \Delta_\mathbf{E}(\mathcal{G}', \mathcal{G}, \lambda) \right] = \lambda \mathbb{E}_{\mathcal{G}'}[a] + (1 - \lambda)\mathbb{E}_{\mathcal{G}'}[d] \\
    &\mathbb{E}_{\mathcal{G}'}[a] = \mathbb{E}_{\mathcal{G}'}\left[\sum_{e \in \mathbf{E}} \mathbbm{1}_{\{e \in E(\mathcal{G'})\}} \cdot (1 - \mathbbm{1}_{\{e \in E(\mathcal{G})\}}) \right]\\
    &\mathbb{E}_{\mathcal{G}'}[d] = \mathbb{E}_{\mathcal{G}'}\left[\sum_{e \in \mathbf{E}} \mathbbm{1}_{\{e \notin E(\mathcal{G'})\}} \cdot \mathbbm{1}_{\{e \in E(\mathcal{G})\}} \right]\\
\end{align}
}

After simplifying the above set of equations and representing $\mathbbm{1}_{\{e \in E(\mathcal{G})\}}$ by the binary variable $x_e$, we arrive at:

\begin{equation}
    \small
    \mathbb{E}_{\mathcal{G}'}\left[ \Delta_\mathbf{E}(\mathcal{G}', \mathcal{G}, \lambda) \right] = \sum_{e \in \mathbf{E}}\left((1 - \lambda) - \mathbb{P}_\mathcal{G'}(e)\right) \cdot x_e + \beta
    \label{eqn:expected_description_length}
\end{equation}
where $\mathbb{P}_\mathcal{G'}(e)$ represents the probability of observing $e \in E(\mathcal{G'})$ when $G' \sim \mathbb{P}_c\left(\cdot, \mathcal{T}\right)$
and $\beta$ is a constant that is independent from the hypothesis $\mathcal{G}$. For each edge $e \in \mathbf{E}$, the desirability of adding $e$ to the hypothesis $\mathcal{G}$ is proportional to the difference between $\mathbb{P}_{\mathcal{G}'}(e)$ and $(1 - \lambda)$. 
As the probability of $e$ exceeds $(1 - \lambda)$ by a larger extent, its desirability to be included in the hypothesis increases as higher probability suggests that the edge $e$ would be present in a significant number of samples, suggesting it is a consistent property. 
Conversely, the presence of $(1 - \lambda)$ in each coefficient prevents the inclusion of edges with probabilities lower than $(1 - \lambda)$ in the aggregated graph.

\subsection{Hypothesis Selection}
\label{sec:hypothesis_selection}
We seek to find $\mathcal{G}$ that minimizes the expected description length in Eq.~\ref{eqn:expected_description_length}. To estimate $\mathbb{P}_\mathcal{G'}(e)$, we compute the fraction of graph samples from $\mathbb{P}_c(\cdot, \mathcal{T})$ that contains $e$ as one of its edges. Formally, we wish to find the following:
\begin{equation}
    \small
    \arg\min_{\mathcal{G}} \sum_{e \in \mathbf{E}}\left((1 - \lambda) - \frac{\sum_{i=1}^T\mathbbm{1}_{e \in E(\mathcal{G}'_i)}}{T}\right) \cdot x_e
    \label{eqn:mdl_simplified}
\end{equation}
In the absence of any additional constraints, identifying the structure becomes trivial---one can simply set $x_e$ to $1$ if its coefficient is negative, and $0$ otherwise. However, in various tasks~\cite{stab2017parsing,saha-etal-2021-explagraphs,sakaguchi-etal-2021-proscript-partially}, the graphs need to be directed acyclic graphs (DAG). Appendix \ref{sec:ilp_dag} explains how to restrict the search space to DAGs when optimizing Eq.~\ref{eqn:mdl_simplified}.

\subsection{Objective for Generic Graphs}
\label{sec:generic_graph_objective_formulation}

Next, we propose suitable modifications to the objective to accommodate generic graphs that may contain singleton nodes. While Eq.~\ref{eqn:dl_no_singleton} defines the description only in terms of edge transformations, we define the description length for generic graphs as follows:

\begin{equation}
    \label{eqn:dl_general}
    \small
    \Delta(\mathcal{G}', \mathcal{G}, \{\lambda_1, \lambda_2\}) = \Delta_\mathbf{E}(\mathcal{G}', \mathcal{G}, \lambda_1) + \Delta_\mathbf{N}(\mathcal{G}', \mathcal{G}, \lambda_2)
\end{equation}
where $\Delta_\mathbf{N}$ uses the same form as Eq.~\ref{eqn:dl_no_singleton} but calculates the description length associated with the addition and removal of nodes in order to transform $N(\mathcal{G})$ into $N(\mathcal{G'})$. Redoing the steps described in \S\ref{sec:eqn_derivation} and \S\ref{sec:hypothesis_selection} yields the following objective:

\begin{equation}
    \small
    \begin{aligned}
        \arg\min_{\mathcal{G}} & \sum_{e \in \mathbf{E}}\left((1 - \lambda_1) - \frac{\sum_{i=1}^T\mathbbm{1}_{e \in E(\mathcal{G}'_i)}}{T}\right) \cdot x_e\\
        + & \sum_{n \in \mathbf{N}}\left((1 - \lambda_2) - \frac{\sum_{i=1}^T\mathbbm{1}_{n \in N(\mathcal{G}'_i)}}{T}\right) \cdot y_n
    \end{aligned}
    \label{eqn:mdl_generic}
\end{equation}
where $y_n$ is a binary variable denoting the presence of $n$ in $N(\mathcal{G})$. Note that, an edge $(n_1, n_2)$ can only exist if both $n_1$ and $n_2$ are present in the graph. To enforce this, we have the constraint: $\forall n_1, n_2 \in \mathbf{N}: y_{n_1} + y_{n_2} - 2x_{(n_1, n_2)} \geq 0$. 

\begin{figure}
    \centering
    \includegraphics[scale=0.37]{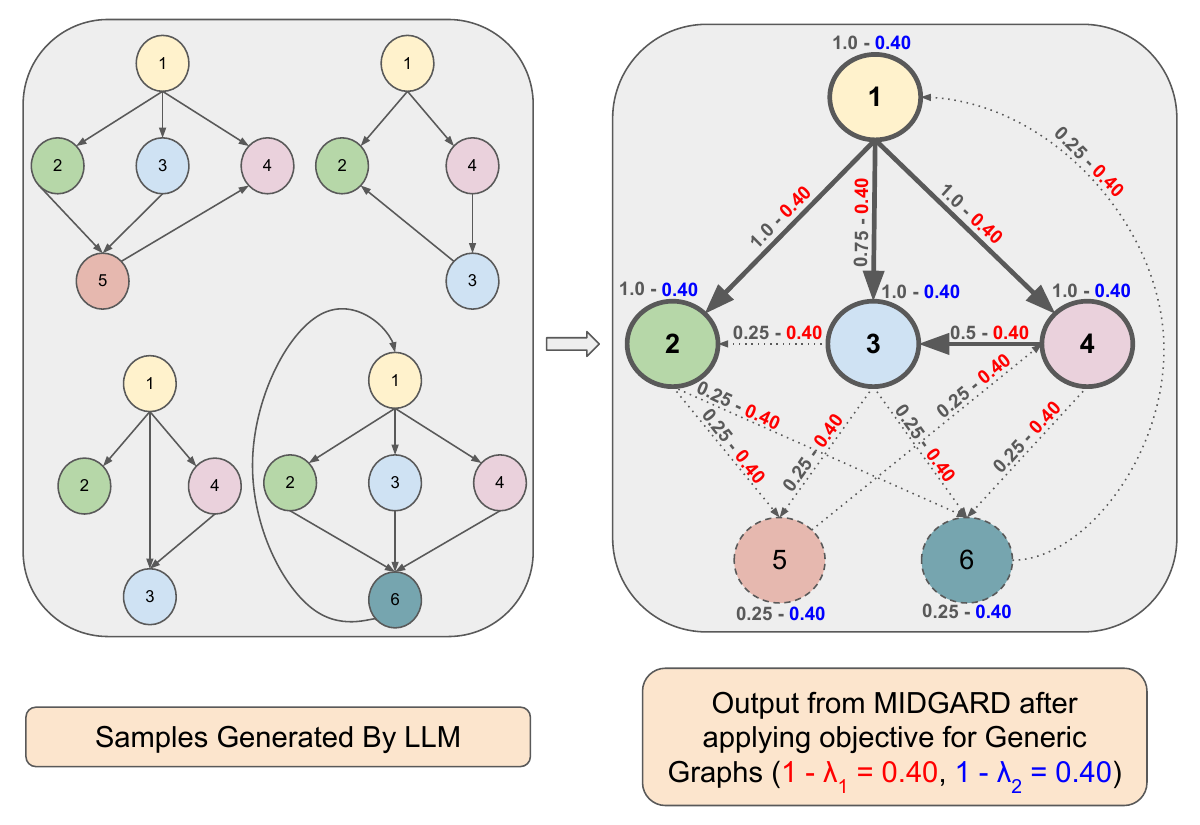}
    \caption{{\bf Pictorial representation of Graph Aggregation.} In the figure above, the probabilities of node/edge existence in a randomly generated sample from an LLM are estimated by the normalized frequency of their occurrence in the samples. The weight of an edge or node on the right-hand side is determined by subtracting $(1 - \lambda_1)$ or $(1 - \lambda_2)$ from this probability, respectively. The optimization in Eq.~\ref{eqn:mdl_generic} is equivalent to the selection of the properties in the aggregated graph such that the sum of weights is maximized. The bolded elements are selected according to this maximization. }
    \label{fig:mdl_explanation}
    
\end{figure}

Refer Figure \ref{fig:mdl_explanation} for pictorial representation of aggregation using the objective in Eq.~\ref{eqn:mdl_generic}.

\section{Experiments and Analysis}

We evaluate on three major tasks for reasoning graph generation: 
\textbf{Task 1-argument structure extraction} on \textsc{Essays}~\cite{stab2017parsing}, \textsc{AbstRCT}~\cite{mayer2020transformer}, and \textsc{CDCP}~\cite{park-cardie-2018-corpus}; 
\textbf{Task 2-generating structured explanations} on \textsc{Explagraphs}~\cite{saha-etal-2021-explagraphs}; 
\textbf{Task 3-script planning} on \textsc{Proscript}~\cite{sakaguchi-etal-2021-proscript-partially}; and \textbf{Task 4-semantic graph generation} on \textsc{Kelm}~\cite{agarwal-etal-2021-knowledge}, \textsc{WebNLG}~\cite{gardent-etal-2017-webnlg}, and \textsc{GenWiki}~\cite{jin-etal-2020-genwiki}. 
Thereafter, we evaluate how the performance of our approach changes when using different numbers of samples generated from the LLM. 
Additionally, we examine the capability of our approach in handling graphs with varied complexities. Finally, we assess the influence of varying the number of few-shot examples and examine how close our automatically chosen hyperparameters are in comparison with the best possible ones.
We also analyze the influence of varying the number of few-shot examples. Unless stated otherwise, we generate $T=10$ samples for approaches utilizing multiple samples. 

\smallskip
\noindent \textbf{Base LLMs.} 
We evaluate our approach with (a) \texttt{gpt-3.5-turbo}\footnote{\url{https://openai.com/chatgpt} (Version \texttt{0613})}, a general purpose instruction-tuned LLM and (b) \textsc{Code-Llama}~\cite{roziere2023code}, a code-LLM pretrained over general purpose programming languages. The $16$K context length associated with these LLMs allows us to employ few-shot prompting for long sequence input-output tasks such as argument structure extraction.

\smallskip
\noindent \textbf{Comparisons.} 
We first consider a \textbf{\textsc{Greedy}} baseline that represents each graph as a semantically equivalent programming script and samples only one generation from the LLM, which is decoded greedily as done in \textsc{CoCoGen}. 

Our main model, \model applies the objective described in \S\ref{sec:generic_graph_objective_formulation}.
As the graphs in all the considered tasks are directed acyclic in nature except semantic graph generation, we additionally incorporate the DAG constraints discussed in \S\ref{sec:hypothesis_selection}. For semantic graph generation, we analyse the performance of different variants without DAG constraints.
We further compare with three variants of \model: 
(a) \textbf{\textsc{\model w/o node trns}}: 
We use the objective described in \S \ref{sec:hypothesis_selection} along with the DAG constraints. By excluding the term that incorporates node transformations in this formulation, we can evaluate its impact on the overall performance. Specifically, this approach is implemented by retaining only those edges that occur more than $1 - \lambda_1$ fraction of times while ensuring there are no cycles. Thereafter, only the nodes present in the retained edges are kept.
(b) \textbf{\textsc{\model ($\lambda=0.5$)}}: 
We apply the objective proposed by \citet{lam1994using} by assuming equal description length of addition and deletion. 
(c) \textbf{\textsc{\model w/o DAG}} constraints.

\begin{table*}[h]
    \centering
    \setlength\tabcolsep{2.5pt}
    \small
    \begin{tabular}{c|ccc|ccc|ccc|ccc}
        \midrule
        \multirow{2}{*}{Approach} & \multicolumn{3}{c}{\textsc{Essays}} & \multicolumn{3}{|c|}{\textsc{AbstRCT}} & \multicolumn{3}{c|}{\textsc{CDCP}} &\multicolumn{3}{c}{Macro Average} \\
        & {\scriptsize C} & {\scriptsize R$_{100}$} & {\scriptsize R$_{50}$ } & {\scriptsize C } & {\scriptsize R$_{100}$ } & {\scriptsize R$_{50}$} & {\scriptsize C} & {\scriptsize R$_{100}$} & {\scriptsize R$_{50}$} & {\scriptsize C} & {\scriptsize R$_{100}$ } & {\scriptsize R$_{50}$ }\\
        \midrule
        \multicolumn{13}{c}{LLM: \texttt{gpt-3.5-turbo}}\\
        \midrule
        \scriptsize{\textsc{Greedy}} & \colorbox[HTML]{c0e7f6}{67.4} & 21.5 & \colorbox[HTML]{c0e7f6}{32.6} & \bf\colorbox[HTML]{d2e7d6}{84.4} & 38.5 & 55.2 & 53.8 & 11.2 & 16.2 & \colorbox[HTML]{c0e7f6}{68.5} & 23.7 & 34.7\\
        \scriptsize{\textsc{\model w/o node trns}} & 65.8 & \bf\colorbox[HTML]{d2e7d6}{23.5} & \bf\colorbox[HTML]{d2e7d6}{35.4} & 83.4 & \bf\colorbox[HTML]{d2e7d6}{41.1} & \bf\colorbox[HTML]{d2e7d6}{58.0} & 48.5 & \bf\colorbox[HTML]{d2e7d6}{12.4} & \bf\colorbox[HTML]{d2e7d6}{18.2} & 65.9 & \bf\colorbox[HTML]{d2e7d6}{25.7} & \bf\colorbox[HTML]{d2e7d6}{37.2}\\
        \scriptsize{\textsc{\model ($\lambda=0.5$)}} & 65.8 & 21.8 & 31.3 & 83.6 & \colorbox[HTML]{c0e7f6}{40.5} & 55.9 & \colorbox[HTML]{c0e7f6}{54.4} & 10.7 & 14.9 & 57.9 & 24.3 & 34.0 \\
        \scriptsize{\textsc{\model w/o DAG}} & \bf\colorbox[HTML]{d2e7d6}{72.3} & \bf\colorbox[HTML]{d2e7d6}{23.5} & \bf\colorbox[HTML]{d2e7d6}{35.4} & \colorbox[HTML]{c0e7f6}{84.0} & \bf\colorbox[HTML]{d2e7d6}{41.1} & \colorbox[HTML]{c0e7f6}{57.9} & \bf\colorbox[HTML]{d2e7d6}{54.8} & \colorbox[HTML]{c0e7f6}{12.3} & \colorbox[HTML]{c0e7f6}{17.9} & \bf\colorbox[HTML]{d2e7d6}{70.4} & \colorbox[HTML]{c0e7f6}{25.6} & \colorbox[HTML]{c0e7f6}{37.1}\\
        \scriptsize{\model} & \bf\colorbox[HTML]{d2e7d6}{72.3} & \bf\colorbox[HTML]{d2e7d6}{23.5} & \bf\colorbox[HTML]{d2e7d6}{35.4} & \colorbox[HTML]{c0e7f6}{84.0} & \bf\colorbox[HTML]{d2e7d6}{41.1} & \bf\colorbox[HTML]{d2e7d6}{58.0} & \bf\colorbox[HTML]{d2e7d6}{54.8} & \bf\colorbox[HTML]{d2e7d6}{12.4} & \bf\colorbox[HTML]{d2e7d6}{18.2} & \bf\colorbox[HTML]{d2e7d6}{70.4} & \bf\colorbox[HTML]{d2e7d6}{25.7} & \bf\colorbox[HTML]{d2e7d6}{37.2} \\
        \midrule
        \multicolumn{13}{c}{LLM: \textsc{Code-Llama}}\\
        \midrule
        \scriptsize{\textsc{Greedy}} & 56.3 & 9.1 & 21.4 & \bf\colorbox[HTML]{d2e7d6}{64.2} & 18.3 & 25.5 & \colorbox[HTML]{c0e7f6}{41.9} & 7.0 & 9.4 & \colorbox[HTML]{c0e7f6}{54.1} & 11.5 & 18.8\\
        \scriptsize{\textsc{\model w/o node trns}} & \colorbox[HTML]{c0e7f6}{56.6} & \bf\colorbox[HTML]{d2e7d6}{11.0} & \bf\colorbox[HTML]{d2e7d6}{24.5} & 57 & \colorbox[HTML]{c0e7f6}{30.7} & \colorbox[HTML]{c0e7f6}{39.5} & 34.5 & \bf\colorbox[HTML]{d2e7d6}{7.8} & \bf\colorbox[HTML]{d2e7d6}{10.7} & 49.4 & \bf\colorbox[HTML]{d2e7d6}{16.5} & \bf\colorbox[HTML]{d2e7d6}{24.9} \\
        \scriptsize{\textsc{\model ($\lambda=0.5$)}} & 49.7 & 3.4 & 5.6 & 62.9 & 24.9 & 31.7 & 36.6 & 3.3 & 4.1 & 49.7 & 10.5 & 13.8 \\
        \scriptsize{\textsc{\model w/o DAG}} & \bf\colorbox[HTML]{d2e7d6}{60.3} & \colorbox[HTML]{c0e7f6}{10.9} & \colorbox[HTML]{c0e7f6}{24.4} & \colorbox[HTML]{c0e7f6}{63.4} & \bf\colorbox[HTML]{d2e7d6}{30.8} & \bf\colorbox[HTML]{d2e7d6}{39.7} & \bf\colorbox[HTML]{d2e7d6}{42.5} & \colorbox[HTML]{c0e7f6}{7.2} & \colorbox[HTML]{c0e7f6}{9.7} & \bf\colorbox[HTML]{d2e7d6}{55.6} & \colorbox[HTML]{c0e7f6}{16.3} & \colorbox[HTML]{c0e7f6}{24.6} \\
        \scriptsize{\model} & \bf\colorbox[HTML]{d2e7d6}{60.3} & \bf\colorbox[HTML]{d2e7d6}{11.0} & \bf\colorbox[HTML]{d2e7d6}{24.5} & \colorbox[HTML]{c0e7f6}{63.4} & \colorbox[HTML]{c0e7f6}{30.7} & \colorbox[HTML]{c0e7f6}{39.5} & \bf\colorbox[HTML]{d2e7d6}{42.5} & \bf\colorbox[HTML]{d2e7d6}{7.8} & \bf\colorbox[HTML]{d2e7d6}{10.7} & \bf\colorbox[HTML]{d2e7d6}{55.6} & \bf\colorbox[HTML]{d2e7d6}{16.5} & \bf\colorbox[HTML]{d2e7d6}{24.9} \\
        \bottomrule
    \end{tabular}
    \caption{
    Results for the argument structure extraction tasks. 
    The results were averaged across 5 random seeds.  
    \colorbox[HTML]{d2e7d6}{\textbf{Green}} 
    and \colorbox[HTML]{c0e7f6}{Blue} indicates best and second-best performance respectively.
}
    \label{tab:argument_structure_extraction}
    \vspace{-10pt}
\end{table*}

\subsection{Task 1: Argument Structure Extraction}
\label{sec:argument_structure_extraction}
In this task, our goal is to analyze the argumentative discourse structure of an input text. This involves detecting and categorizing all argumentative components within the text and identifying the relationships between them. An example of this task is shown in Appendix \ref{sec:appendix_argument_structure}. 

We assess the performance of our method on the following datasets: \textbf{(1) \textsc{Essays}}~\cite{stab2017parsing},
\textbf{(2) \textsc{AbstRCT}}~\cite{mayer2020transformer}, 
\textbf{(3) \textsc{CDCP}}~\cite{park-cardie-2018-corpus}. For more details on these details, please refer the appendix \ref{sec:appendix_argument_structure}.

To compute the performance of component identification, we use BIO scheme to label the token sequences. Thereafter, we compute component identification $F_1$ score (\textbf{C}) by tallying the number of true positives (TP), false negatives (FN), and false positives (FP) in the assigned token sequences as specified by \citep{mayer2020transformer}. To assess the performance of relation prediction, we compute metrics denoted by \textbf{R$_{100}$} and \textbf{R$_{50}$}. The R$_{100}$ metric computes the $F_1$-score by considering a prediction as true positive only if the head and tail components (and the relation type) overlap exactly with that of a ground truth edge. On the other hand, R$_{50}$ considers a predicted relation as correct if there is at least a $50\%$ token overlap between the head and tail components of the prediction and a ground-truth relation.

We observe from Table \ref{tab:argument_structure_extraction} that \textit{\model achieves a consistent performance improvement across component identification and relation prediction} for most of the datasets and LLM choices. With just $10$ samples, the component identification performance of \textsc{Essays} is elevated by $\approx 5\%$ and $\approx 4\%$, using $\texttt{gpt-3.5-turbo}$ and $\textsc{Code-Llama}$ respectively. Our approach boosts the performance for \textsc{AbstRCT} for relation prediction by over $10\%$ when \textsc{Code-Llama} is used.  

\textsc{\model} consistently outperforms other aggregation strategies, with \textsc{\model w/o DAG} being only slightly inferior. This indicates that \textsc{\model w/o DAG} can be used for argument mining tasks without a significant decline in performance, even without incorporating DAG constraints for graph combination. However, when it comes to component identification, \textsc{\model w/o node trns} yields poor results due to the absence of the term describing node transformations. Table \ref{tab:argument_structure_extraction} also justifies why it is important to have unequal description lengths for addition and deletion. 

In our analysis of the models' errors, we observed that LLMs excel in accurately identifying specific components. However, they tend to miss capturing all the components present in the data. On the other hand, \model effectively assimilates relevant components from multiple samples, resulting in improved recall without compromising precision (refer Appendix~\ref{sec:precision_recall_argument} for quantitative results). 

Furthermore, as shown in Table \ref{tab:error_categories}, our approach is effective in \textbf{reducing various types of errors} found in the inferred edges. We categorize errors associated with relation prediction and calculate the total count of distinct edge errors in the inferred samples as compared to the ground truth data.

\begin{table}
    \scriptsize
    \setlength\tabcolsep{2.5pt}
    \centering
    \begin{tabular}{c|cc|cc|cc}
        \midrule
        \multirow{2}{*}{Type of error} & \multicolumn{2}{c|}{\textsc{Essays}} & \multicolumn{2}{c|}{\textsc{AbstRCT}} & \multicolumn{2}{c}{\textsc{CDCP}} \\
        & \small Grdy & \small Ours &  \small Grdy & \small Ours & \small Grdy & \small Ours \\
        \midrule
        \# spurious edges included & 528.2 & \bf469.0 & 200.2 &	\bf182.0	& 441.6 &	\bf411.2 \\
        \# true edges omitted & 777.8 & \bf768.0 & 126.8 &	\bf120.2 &	255.6	& \bf238.8\\
        \# reversed edges& 30.4 & \bf21.8 & 1.4 & \bf0.8 & 15.8 & \bf14.4\\
        \bottomrule
    \end{tabular}
    \caption{Segregation of relation prediction errors into distinct categories and assessing how effective \model (Ours) is reducing various types of errors more than \textsc{Greedy} (Grdy) decoding. Performance averaged over $5$ random seeds.}
    \label{tab:error_categories}
\end{table}

\subsection{Task 2: Explanation Graph Generation}
\label{sec:explanation_graph_generation}
\begin{table}[t]
    \scriptsize
    \centering
    \setlength\tabcolsep{1.5pt}
    \begin{tabular}{c|cccc}
        \midrule
        \multirow{2}{*}{Approach} & \multicolumn{4}{c}{\textsc{Explagraph}} \\
        & StCA ($\uparrow$) & SeCA ($\uparrow$) & G-BS ($\uparrow$) & GED ($\downarrow$)\\
        \midrule
        \multicolumn{5}{c}{LLM: \texttt{gpt-3.5-turbo}} \\
        \midrule
        {\textsc{Greedy}} & 23.7 & 7.6 & 18.6 & 84.0 \\
        {\textsc{\model w/o DAG}} & 12.9 & 2.5 & 10.3 & 91.1\\
        {\textsc{\model} ($\lambda = 0.5$)} & 4.3 & 1.6 & 3.3 & 97.0\\
        {\textsc{\model}} & \bf 30.3 & \bf 17.7 & \bf 22.4 & \bf 82.1\\
        \midrule
        \multicolumn{5}{c}{LLM: \textsc{Code-Llama}} \\
        \midrule
        {\textsc{Greedy}} & 36.6 & 12.4 & 28.4 & \bf 75.6 \\
        {\textsc{\model w/o DAG}} & 21.2 & 12.6 & 16.1 & 87.3 \\
        {\textsc{\model} ($\lambda = 0.5$)} & 0.0 & 0.0 & 0.0 & 100.0 \\
        {\textsc{\model}} & \bf 39.4 & \bf 20.2 & \bf 29.7 & 76.4 \\
        \bottomrule
    \end{tabular}
    \caption{
    Results on \textsc{Explagraph}.
    \model ($\lambda = 0.5$) resulted in none of the edges being included in the final graph, as the estimated probabilities of all edges in the samples are below $0.5$. 
}
    \label{tab:explanation_graph_generation}
\end{table}

We use \textsc{Explagraphs}~\cite{saha-etal-2021-explagraphs} for this task, where the goal is to predict whether a certain argument supports or counters a belief while generating a commonsense explanation graph that explicitly conveys the reasoning behind the stance prediction. We request the reader to refer Appendix~\ref{sec:appendix_explanation_graph_generation} for more details on prompt design and dataset.

We employ the following metrics recommended by the authors of this task~\cite{saha-etal-2021-explagraphs}: 
(1) Structural Accuracy (\textbf{StCA}) computes fraction of graphs that are DAG and has $2$ concepts from the argument and belief. 
(2) Semantic Correctness (\textbf{SeCA}) employs a learnt model to measure the semantic correctness of the edges by checking whether the implied stance from the graph matches the ground truth.  
(3) G-BERTScore (\textbf{G-BS}) measures the BERTScore~\cite{zhang2019bertscore} between the inferred and ground truth edges. 
(4) Graph Edit Distance (\textbf{GED}) computes graph edits required to transform the hypothesis to the ground truth. As these evaluation metrics measure the accuracy of the graph as a collection of edges, we do not experiment with \textsc{\model w/o node trns} as there would be no performance difference from \model.

We observe from Table \ref{tab:explanation_graph_generation} that  \textsc{\model} \textit{improves the performance of single generation based technique by a significant margin} for both LLMs. Unlike the argument structure extraction task, the performance is significantly worse when not using the DAG constraints. 

\subsection{Task 3: Script Planning}

Unlike the previous two tasks which emphasize on constructing the complete graph from scratch, we investigate whether our approach can be used for inferring relations between nodes that are already known. To examine this, we use \textsc{Proscript}~\cite{sakaguchi-etal-2021-proscript-partially}, which involves generating a graph for achieving a high-level goal, with each node representing an action and edges indicating dependency relations among actions. In our setup, we provide the set of actions and the goal to the LLM as input and prompt it to generate the sequence of edges that capture the dependencies among the input actions. More details about this task is presented in Appendix~\ref{sec:appendix_script_planning}.

\begin{figure}
    \centering
    \includegraphics[scale=0.25]{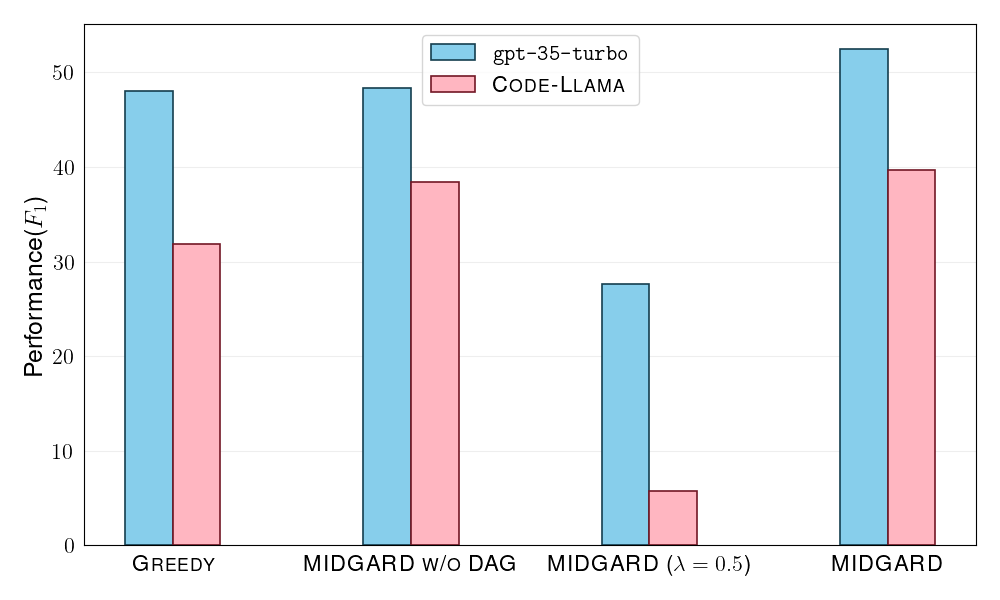}
    \vspace{-3mm}
    \caption{Results for script planning on \textsc{Proscript}. 
    }
    \label{fig:script_planning}
    \vspace{-3mm}
\end{figure}

To compare the performance between different approaches, we use F1-score ($F_1$) between the inferred edge set and the ground truth. From Figure~\ref{fig:script_planning}, we can see that \model significantly improves the performance over the greedy single-generation based approach. The figure also demonstrates the importance of having DAG constraints. 

\subsection{Task 4: Semantic Graph Generation}
\begin{table*}[h]
    \centering
    \scriptsize
    \setlength\tabcolsep{2.5pt}
    \begin{tabular}{c|cccc|cccc|cccc}
        \midrule
        \multirow{2}{*}{Approach} & \multicolumn{4}{|c|}{\textsc{Kelm}} & \multicolumn{4}{c}{\textsc{WebNLG}} & \multicolumn{4}{|c}{\textsc{GenWiki}} \\
        & T-F$_1$($\uparrow$) & G-F$_1$($\uparrow$) & G-BS($\uparrow$) & GED($\downarrow$) & T-F$_1$($\uparrow$) & G-F$_1$($\uparrow$) & G-BS($\uparrow$) & GED($\downarrow$) & T-F$_1$($\uparrow$) & G-F$_1$($\uparrow$) & G-BS($\uparrow$) & GED($\downarrow$)\\
        \midrule
        \multicolumn{13}{c}{LLM: \texttt{gpt-3.5-turbo}} \\
        \midrule
        \textsc{Greedy} &  46.9 & \bf 22.8 & \bf 84.0 & \bf 8.7 & 29.1 & \bf 15.0 & 83.6 & \bf 10.4 & 23.7 & 6.5 & 82.5 & 11.9\\
        \model $(\lambda = 0.5)$ &  47.0 & 22.0 & 83.2 & 8.9 &  27.8 & 13.2 & 82.4 & 10.7 &  24.0 & 7.0 & 82.4 & 11.6 \\
        \model &  \bf 47.4 & \bf 22.8 & 83.5 & 8.8 & \bf 29.3 & \bf 15.0 & \bf 83.7 & \bf 10.4 & \bf 24.3 & \bf 7.2 & \bf 83.4 & \bf 11.5\\
        \midrule
        \multicolumn{13}{c}{LLM: \textsc{Code-Llama}} \\
        \midrule
        \textsc{Greedy} & \bf 37.9 & \bf 20.0 & 63.2 & 14.1 & 24.8 & \bf 6.0 & 66.4 & 14.2 & \bf 12.1 & 2.0 & 53.6 & 17.6\\
        \model $(\lambda = 0.5)$ & 8.8 & 4.0 & 45.0 & 19.8 & 23.0 & \bf 6.0 & 67.3 & 14.6 & 7.1 & 2.0 & 54.9 & 18.2\\
        \model & \bf 37.9 & 12.0 & \bf 67.7 & \bf 13.5 & \bf 26.5 & \bf 6.0 & \bf 77.7 & \bf 12.2 & 9.7 & \bf 4.0 & \bf 58.7 & \bf 17.3 \\
        \midrule
    \end{tabular}
    \caption{Results for Semantic Graph Generation. In each of our method variants, we did not apply \textsc{Dag} constraints, as they are not necessary for this task unlike the previous experiments. 
    }
    \label{tab:semantic_graph_generation}
    \vspace{-10pt}
\end{table*}

The goal of this task is to extract the semantic graph from an input natural language text  as a list of edges.
Each edge in the graph consists of a subject, a property, and the type of property~\cite{han2023pive}. An example of this task is shown in Appendix \ref{sec:additional_semantic_graph_construction}.

To gauge the efficacy of our model for such a task, we consider following datasets: \textbf{(1) \textsc{Kelm}}~\cite{agarwal-etal-2021-knowledge}, \textbf{(2) \textsc{WebNLG}}~\cite{gardent-etal-2017-webnlg} and \textbf{(3) \textsc{GenWiki}}~\cite{jin-etal-2020-genwiki}. For more details on these datasets, we request the reader to refer Appendix \ref{sec:additional_semantic_graph_construction}

We use the following metrics to assess the quantitative performance as suggested by \citet{han2023pive}: \textbf{(1)} Triple-Match F$_1$ (\textbf{T-F$_1$}) finds the macro-averaged F$_1$ between the edge triples present in the inference and the ground truth graph edge triples. \textbf{(2)} Graph Match F$_1$ (\textbf{G-F$_1$}) measures the performance as the number of graphs which exactly matches the ground truth graph in terms of F$_1$ score. Finally, as defined in the \S \ref{sec:explanation_graph_generation}, we also use \textbf{(3)} G-BERTScore (\textbf{G-BS})~\cite{zhang2019bertscore} and \textbf{(4)} Graph Edit Distance (\textbf{GED}).

From the Table \ref{tab:semantic_graph_generation}, we can observe that while our approach outperforms or achieves competitive performance compared to the baseline, the performance improvement is not significant for \texttt{gpt-3.5-turbo}. Upon closer examination of the outputs generated using temperature sampling, we have noticed a lack of variability when compared to the structured commonsense reasoning tasks mentioned in the main script. This limited variability hinders the opportunity to improve upon each sample, resulting in a less significant performance boost than expected.

\subsection{Further Analyses}

\noindent \textbf{Impact of increasing sample size.} 
To analyze the impact of varying the number of samples generated from the LLM, we evaluate the performance of \model on the argument structure extraction task as it would allow us to examine the trend on both node identification and edge prediction. We only show the analysis for the \textsc{Essays} dataset from the argument structure extraction task due to limited space. Please refer to the appendix for additional plots and similar analysis.

\begin{figure}[h]
  \centering  
  \begin{subfigure}[b]{0.23\textwidth}  
    \includegraphics[width=\textwidth]{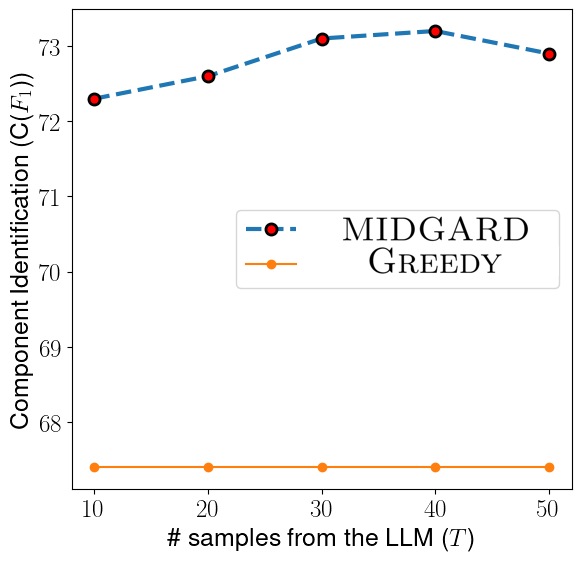}  
    \caption{Component Identification }  
    \label{fig:essay_component_identification}  
  \end{subfigure}  
  \hfill  
  \begin{subfigure}[b]{0.23\textwidth}  
    \includegraphics[width=\textwidth]{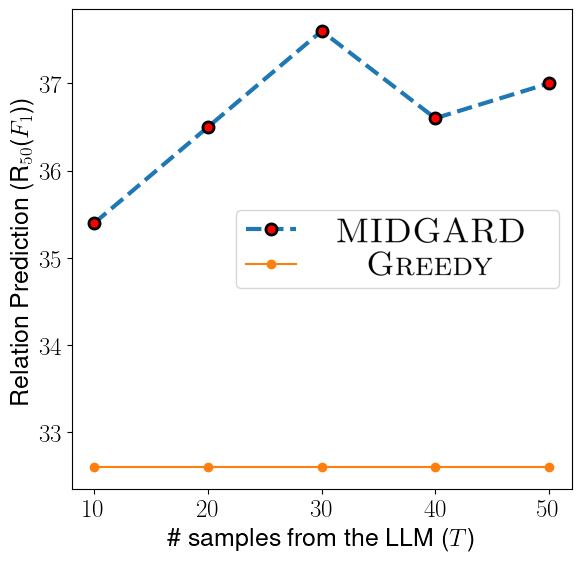}  
    \caption{Relation Prediction}  
    \label{fig:essay_relation_prediction}  
  \end{subfigure}  
  \caption{Performance of \textsc{\model} in comparison with \textsc{Greedy} on \textsc{Essays} when the number of samples from the LLM is varied. Results averaged over $5$ different random seeds.}  
  \label{fig:essay_performance}  
\end{figure}

\begin{figure}[h]
    \centering
    \includegraphics[scale=0.35]{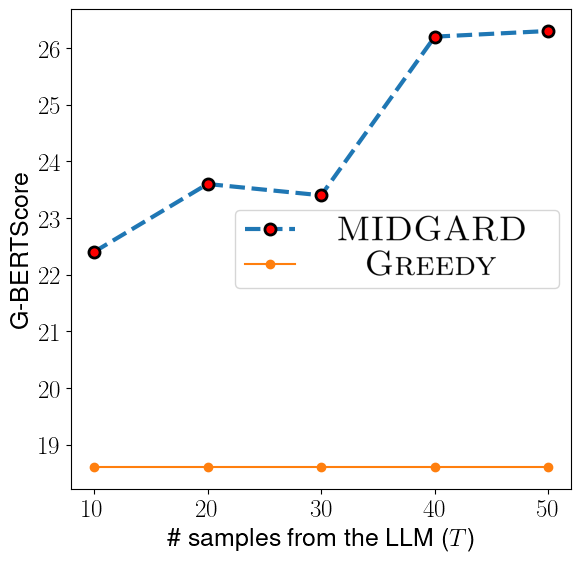}
    \caption{Performance of \textsc{\model} in comparison with \textsc{Greedy} on \textsc{Explagraphs}.}
    \label{fig:explagraph_samples}
\end{figure}

From Fig.~\ref{fig:essay_component_identification} and Fig.~\ref{fig:essay_relation_prediction}, we see that the performance increases only marginally emphasizing that returns diminish with increasing the number of samples. Similar trend is observed for other datasets belonging to the same task (refer Appendix \ref{sec:argument_extraction_samples}). However, for \textsc{Explagraphs}, we observe that the performance steadily increases with the number of samples indicating that having more and diverse explanation graphs is helpful towards improving the final aggregated structure as shown in Figure~\ref{fig:explagraph_samples}.

\label{sec:appendix_explagraph_samples}


\begin{table*}[h]
\centering
\small
\setlength\tabcolsep{3.5pt} 
    \begin{tabular}{ccccc|cc|cc}
        \midrule
        \multirow{2}{*}{Bin} & \multirow{2}{*}{\# Samples} & \multirow{2}{*}{Avg. \# Nodes} & \multirow{2}{*}{Avg. \# Edges} & \multirow{2}{*}{Avg. Degree} & \multicolumn{2}{c|}{\textsc{Greedy}} & \multicolumn{2}{c}{\model}\\
        & & & & & C & R$_{50}$ & C & R$_{50}$ \\
        \midrule
        \multicolumn{9}{c}{\textsc{Essays}}\\
        \midrule
        $[5, 10)$ & 3 & 8.0 & 7.0 & 0.88 & 64.7 & 23.8 & 64.7 & \bf 36.2\\
        $[10, 15)$ &	28 &	11.2 &	10.2 &	0.91 &	66.2 &	33.5 &	\bf 70.2	 & \bf 34.8\\
        $[15, 20)$ &	33	& 15.6 &	14.6 &	0.94 &	68.5	& 32.5 & \bf	73.0 & \bf 	36.4\\
        $[20, 25)$	& 14	& 20.6 &	19.6 &	0.95 &	65.7 &	32.0	& \bf 75.5	& \bf 35.5\\
        $[25, 30)$ &	2 &	26.0 &	25.0 &	0.96 &	58.3 & \bf	36.1 & \bf	70.0 &	31.2\\
        \midrule
        \multicolumn{9}{c}{\textsc{AbstRCT}}\\
        \midrule
        $[2, 4)$	& 5 &	2.8	& 1.4 &	0.47 &	\bf79.4 &	58.2 &	77.6 &	\bf 59.6\\
        $[4, 6)$ &	41 &	4.6 &	2.7 &	0.58 & 83.4 & 	59.8 & \bf	83.5 & \bf	63.1\\
        $[6, 8)$ &	36	& 6.5 &	3.6 &	0.56 & \bf 88.6 &	57.9 &	87.6 & \bf	59.9 \\
        $[8, 10)$ &	14 &	8.4 &	4.0 &	0.47 & 83.8 &	47.0 &	\bf 84.5 &	\bf 48.3 \\
        $[10, 12)$ &	3 &	10.3 &	7.0 &	0.68 & \bf 70.1 &	39.6 &	68.5 &	\bf 50.5 \\
        \midrule
        \multicolumn{9}{c}{\textsc{CDCP}}\\
        \midrule
        $[2, 7)$ &	96 &	3.9 &	1.2 &	0.26 & 52.1 &	20.2	& \bf 52.4 &	\bf 22.3 \\
        $[7, 12)$ &	33 &	8.3 &	3.2 &	0.39 & 56.0 & 	21.7 & \bf	56.9 &	\bf23.8\\
        $[12, 17)$ &	13 &	13.9 &	3.9 &	0.27 & \bf 55.7 &	8.1 &	54.9 &	\bf11.5 \\
        $[17, 22)$ &	3 &	19.0 &	7.7	& 0.40 & 59.7 &	\bf 10.4 &	\bf 64.4 &	7.7 \\
        $[22, 27)$ &	2 &	23.0 &	4.5 &	0.20 & 52.6 &	\bf 1.1 &	\bf 55.8 &	1.0 \\
        \bottomrule
    \end{tabular}
    \caption{
    Component and Relation identification performance for \textsc{Greedy} and \model for different graph complexities when \texttt{gpt-3.5-turbo} is used. The results are averaged for $5$ seeds.}
    \label{tab:complexity_analysis}
\end{table*}

\noindent \textbf{Efficacy of \model for different graph complexities.}
We compare the performance between \model and the \textsc{Greedy} approach on argument structure extraction across various graph complexities. We specifically select argument structure extraction for this analysis because it enables us to evaluate the influence of graph complexity on both node and edge identification performance. 

In this analysis, we bin the graphs based on the number of nodes and compute the average complexity metrics such as number of nodes and edges and degree for the graphs belonging to each bin. The higher these metrics are, the more complex the corresponding graph is. For each method, we employ \texttt{gpt-3.5-turbo} for generating samples. We observe that our approach provided consistent improvements across different complexities as shown in Table \ref{tab:complexity_analysis}.

\noindent\textbf{Additional analysis.} The impact of varying the number of few-shot examples on argument structure extraction performance for \textsc{Greedy} and \model is provided in Appendix \ref{sec:argument_extraction_few_shot}. \model consistently improves the performance across different number of few-shot examples. We compare our method and \textsc{Greedy} against a popular decoding technique called \textsc{Nucleus} Sampling~\cite{Holtzman2020The} in the Appendix \ref{sec:appendix_nucleus_sampling} and find that it results in poorer performance. We demonstrate that our approach works with \texttt{gpt-4} for the \textsc{Essays} dataset in Appendix \ref{sec:appendix_gpt4_evaluation}.
In Appendix \ref{sec:hyperparameter_analysis}, we assess the impact of varying the hyperparameters $\{\lambda_1, \lambda_2\}$ on the final performance and compare it with that of automatically estimated hyperparameters (refer Appendix \ref{sec:hyperparameter_selection}). Figure \ref{fig:essay_threshold_analysis} shows that while our automatic hyperparameter search reaches near optimal performance for component identification, there is a scope for improvement in relation prediction.

\section{Related Works}


\noindent \textbf{Sampling based approaches using LLMs}.
A common strategy to address many NLP and commonsense reasoning tasks involves sampling multiple solution trajectories LLMs and employing either a post-hoc strategy~\cite{fu2023gptscore,liu2023gpteval,wang2023chatgpt} or a trained reranker for sample selection~\cite{cobbe2021training,li-etal-2023-making,ni2023lever}. 
However, post-hoc approaches relying on LLM evaluation can be prone to position bias~\cite{wang2023chatgpt,zheng2023large} and difficulty in judging response correctness~\cite{huang2023large,gou2023critic}. Training-based sampling requires additional labeled data for task-specific reranking models. The \textit{self-consistency} framework is limited to problems with scalar answer spaces due to its reliance on majority voting~\cite{ling-etal-2017-program,clark2018think,cobbe2021training,patel-etal-2021-nlp,geva2021did}. Moreover, existing approaches lack integration of information from different samples, potentially leading to suboptimal solutions. In contrast, our MDL-based formulation assimilates relevant information from diverse structured responses without fine-tuning. By examining consistent properties across samples, we construct an aggregate graph that leverages the strengths of each sample.

\smallskip
\noindent \textbf{LLMs for commonsense reasoning.}
LLMs have been applied to various domains, including arithmetic reasoning~\cite{he2023solving}, generation of mathematical proofs~\cite{welleck2022naturalprover}, symbolic reasoning~\cite{wei2022chain}, and logical reasoning~\cite{srivastava2022beyond}. While prompting strategies~\cite{wei2022chain,zhou2022least,yao2022react,wang2023selfconsistency,yao2023tree,madaan2023self} have been proposed to improve performance across these tasks, adapting them to structured commonsense reasoning, which involves generating complex graph structures, presents unique challenges~\cite{madaan-etal-2022-language}. Additionally, tasks within structured commonsense reasoning often require adherence to specific constraints~\cite{saha-etal-2021-explagraphs,sakaguchi-etal-2021-proscript-partially}, such as directed acyclicity, which are difficult to ensure solely through existing strategies. Our approach is independent of the prompting methodology and allows for flexible incorporation of task-specific constraints during inference.

\section{Conclusion}
We proposed a novel approach for enhancing the performance of structured reasoning problems which involve generating task-specific graphs. 
Taking inspiration from self-consistency, we sample multiple graphs from the LLM and devise a mechanism to construct aggregated graph. Through rigorous experimentation, we have demonstrated the effectiveness of our approach across various structured commonsense reasoning tasks. 

\section*{Limitations}
\begin{itemize}
    \item Due to our approach's reliance on generating multiple samples, it can be computationally demanding and may require a significant amount of time, particularly without batched inference. As a result, practitioners using enterprise LLMs may incur substantially higher costs compared to methods that involve single generation. This factor makes our approach less desirable in situations where there are constraints on compute budget or limited machinery resources.
    \item For datasets consisting of graphs with a small number of nodes and edges, applying ILP does not result in significant computational overhead. However, it is important to acknowledge that the time complexity of ILP solvers grows exponentially with the complexity of the problem. Therefore, modifications are necessary when applying our approach to settings with a large number of edges and nodes. Additionally, as the graph size increases, it becomes increasingly challenging to utilize LLMs effectively in generating the graph structure. The limited context length of the LLMs poses a challenge for applying them to commonsense reasoning tasks involving larger graphs. This limitation arises from the difficulty of accommodating multiple in-context learning examples within the given context length.
\end{itemize}

\section*{Ethics Statement}
While our methodology attempts to derive structured representations from the input data only, due to the issue of hallucination, the LLMs are not immune to generating biased, insensitive or untruthful content. Hence, we urge practitioners and researchers to exercise caution when applying our framework, especially for sensitive applications like politics, finance, and healthcare.

\section*{Acknowledgements}
This work is supported in part through National Science Foundation under grant 2302564. We are grateful for the resources and services provided by Advanced Research Computing (ARC), a division of Information and Technology Services (ITS) at the University of Michigan, Ann Arbor. 
Additionally, we thank the members of the LAUNCH group at the University of Michigan for their discussions and suggestions.

\bibliography{anthology,custom}
\bibliographystyle{acl_natbib}
\bibliographystyle{acl_natbib}

\appendix
\section{Minimum Description Length Principle}
\label{sec:mdl_principle}
The principle of Minimum Description Length (MDL) aims to find a model that can efficiently represent a dataset using the fewest bits, while also minimizing model complexity. In simpler terms, it seeks to find the least complex model that can effectively capture the regularities in a given dataset using the least amount of bits. Let's denote the dataset that needs to be represented as $\mathcal{D}$, and the model as $H \in \mathcal{H}$. We represent the description length of $\mathcal{D}$ when using $H$ as $L(\mathcal{D} | H)$, which quantifies the number of bits required to describe $\mathcal{D}$ using $H$. Additionally, let's define $L(H)$ as the complexity of the model. Formally, MDL aims to find the optimal solution for:
\begin{equation}
    H^{*} = \arg\min_{H \in \mathcal{H}} L(\mathcal{D} | H) + L(H)
\end{equation}

To explain what each of these terms corresponds to in our approach, let's consider the objective for graphs that do not have singleton nodes in \S \ref{sec:hypothesis_selection}, while adhering to the constraint that the sought-after graph is a Directed Acyclic Graph (DAG). In this scenario, the hypothesis family $\mathcal{H}$ encompasses all graphs with nodes and edges in $\mathbf{N}$ and $\mathbf{E}$, respectively. The dataset in our case consists of samples derived from the LLM, which can be denoted as $\mathcal{D} = \{\mathcal{G}'\}_{i=1}^T$. The formulation of $L(\mathcal{D}|H)$ takes the form of Equation \ref{eqn:dl_no_singleton}. Lastly, we define the complexity of the model $L(H)$ as $0$ if $H$ is a DAG, and $\infty$ otherwise.

\section{Restricting hypothesis selection to DAGs}
\label{sec:ilp_dag}

To describe the strategy to restrict hypothesis selection to DAGs, we can express $\mathbf{E}$ as the set $\{(n_1, n_2) \, | \, n_1 \in \mathbf{N}, n_2 \in \mathbf{N}\}$ where $n_1$($n_2$) represents the head (tail) of the edge $(n_1, n_2)$. We formulate objective \ref{eqn:mdl_simplified} as an Integer Linear Programming (ILP) problem by introducing one more binary variable $b_e$ for each edge $e \in \mathbf{E}$. $b_e$ is set as $1$ if there exists a path from the head of $e$ to its tail. Under ILP, we optimize \ref{eqn:mdl_simplified} subject to the following constraints:

\begin{align}
  \phantom{\forall n_1, n_2, n_3 \in \mathbf{N}}
  &\begin{aligned}
    \mathllap{\forall e \in \mathbf{E}} &: x_e - b_e \leq 0\\
    \label{eqn:direct_edge}
  \end{aligned}\\
  &\begin{aligned}
    \mathllap{\forall n_1, n_2, n_3 \in \mathbf{N}} &: b_{(n_1, n_3)} - b_{(n_1, n_2)} \\
      & - b_{(n_2, n_3)} \geq -1\\
    \label{eqn:transitivity}
  \end{aligned}\\
  &\begin{aligned}
    \mathllap{\forall n \in \mathbf{N}} &: b_{(n, n)} = 0\\
    \label{eqn:no_cycle}
  \end{aligned}
\end{align}

The constraint represented by \ref{eqn:direct_edge} ensures that there is a path between two nodes if they are directly connected by an edge. \ref{eqn:transitivity} enforces the requirement that a path must exist between two nodes if there is a path from the first node to a third node, and this third node is connected to the second node. Lastly, \ref{eqn:no_cycle} prevents any cycles from occurring in the graph.

\section{Implementation Details}
\label{sec:implementation_details}
In this section, we begin by explaining the construction of $\mathbf{N}$ and $\mathbf{E}$ based on the samples $\{\mathcal{G'}_i\}_{i=1}^T \sim \mathbb{P}_c\left(\cdot, \mathcal{T}\right)$. Thereafter, we describe how the hyperparameters $\lambda_1$ and $\lambda_2$ are set. 

\subsection{Constructing $\mathbf{N}$ and $\mathbf{E}$}
To build $\mathbf{N}$, we iterate through the samples $\{\mathcal{G}'_i\}_{i=1}^T$. It is 
important to note that each node in $\mathcal{G}'_i$ consists of two primary properties: \texttt{content} and \texttt{type}. For example, in argument structure extraction~\cite{stab2017parsing}, a node represents an argumentative component with \texttt{content} indicating its value and \texttt{type} indicating its category, such as premise/claim.

When we are iterating over the nodes in $\bigcup_{i=1}^TN(\mathcal{G'}_i)$, we have two choices: append it as a new node or merge it with some other node present in $\mathbf{N}$. To keep track of the historical merging of nodes with $n \in \mathbf{N}$, we maintain two lists: \texttt{content\_list} and \texttt{type\_list}. These lists store the \texttt{content} and \texttt{type} properties of the nodes that have been merged with $n$ over time, respectively. Moreover, \texttt{content\_list} property can also be used to decide whether a new node has to merged. If the Jaccard similarity between the set of tokens in the \texttt{content} of the new node and the set of tokens in an element of \texttt{content\_list} for $n$ exceeds a pre-defined threshold, we add the \texttt{content} and \texttt{type} properties of the new node to the respective lists associated with $n$. Finally, the sentence from the \texttt{content\_list} with the highest Jaccard similarity to the rest of the elements, and the mode of the \texttt{type\_list} of $n$, are chosen as its \texttt{content} and \texttt{type}, respectively. The number of samples containing $n$ is simply the length of its \texttt{content\_list} which can be used to estimate $\mathbb{P}_{\mathcal{G}'}(n)$.

Similarly, we initialize $\mathbf{E} = \{(n_1, n_2) \,|\, n_1, n_2 \in \mathbf{N}\}$. Just like before, each edge in any sample is linked to a specific \texttt{type} property that characterizes the attribute associated with it. For example, in argument structure extraction, the \texttt{type} of an edge can be defined as attack or support, indicating the relationship between the head and the tail of the edge. As before, we associate \texttt{type\_list} property to each $e \in \mathbf{E}$, which records the observed type property for that edge across all the samples it appears in. Finally, the \texttt{type} of each edge is the mode of its \texttt{type\_list} property.

\subsection{Constructing Optimal Aggregate Graph}
After constructing $\mathbf{N}$ and $\mathbf{E}$, we apply an appropriate formulation of the objective in \S \ref{sec:mdl_for_objective_formulation} to get the optimal values of $x_e (\forall e \in \mathbf{E})$ and $y_n (\forall n \in \mathbf{N})$. Thereafter, we return the hypothesis $\mathbf{G}$ where $N(\mathcal{G}) = \{n \,|\, y_n = 1, n \in \mathbf{N}\}$ and $E(\mathcal{G}) = \{e \,|\, x_e = 1, e \in \mathbf{E}\}$. If singleton nodes are absent, we only retain nodes that are present as a head or tail in $E(\mathcal{G})$. 

\subsection{Hyperparameter selection}
\label{sec:hyperparameter_selection}
To automatically select appropriate values for the hyperparameters $\{\lambda_1, \lambda_2\}$, we utilize k-fold cross validation using the few-shot examples. In each fold, the held-out set comprises a single data point, while the training set consists of $k-1$ data points.

\subsection{Generating graphs from LLMs}
\label{sec:generating_graph_from_LLM}
For each dataset, the graph structure is encoded as a programming script following the guidelines of \textsc{CoCoGen} (refer appendix \ref{sec:appendix_argument_structure}). Multiple samples are generated from the LLM using a temperature of $0.9$. To address the randomness in sampling few-shot examples and temperature sampling, we use 5 different random seeds.

Once the LLM generates the graph as a programming script, we obtain the corresponding graph $\mathcal{G}'$, a parser is needed to process this output. During sampling from the LLM, we assume that the textual response can be parsed into the corresponding graph using a task-specific rule-based parser. 



\section{Additional information on considered tasks}

\subsection{Task 1: Argument Structure Extraction - Additional Information on task, prompt design and datasets}
\label{sec:appendix_argument_structure}
In this section, we show an example of this task and describe the prompt used for our experiments. The example for demonstrating this task is taken from one of the paragraphs in a datapoint belonging to the \textsc{Essays} dataset.

\begin{table}[h]  
\centering  
\begin{tabular}{p{0.8\linewidth}} 
\midrule  
First, [\uline{cloning will be beneficial for many people who are in need of organ transplants}]$_{\textrm{Claim 1}}$. [\uwave{Cloned organs will match perfectly to the blood group and tissue of patients}]$_{\textrm{Premise 1}}$ since [\uwave{they can be raised from cloned stem cells of the patients}]$_{\textrm{Premise 2}}$. In addition, [\uwave{it shortens the healing process}]$_{\textrm{Premise 3}}$. Usually [\uwave{it is very rare to find an appropriate organ donor}]$_{\textrm{Premise 4}}$ and [\uwave{by using cloning in order to raise required organs the waiting time can be shortened tremendously}]$_{\textrm{Premise 5}}$.\\  
\bottomrule  
\end{tabular}
\caption{An example text with annotated argumentative components.}
\label{tab:argument_text_annotated}
\end{table} 

\begin{figure}
    \centering
    \includegraphics[scale=0.3]{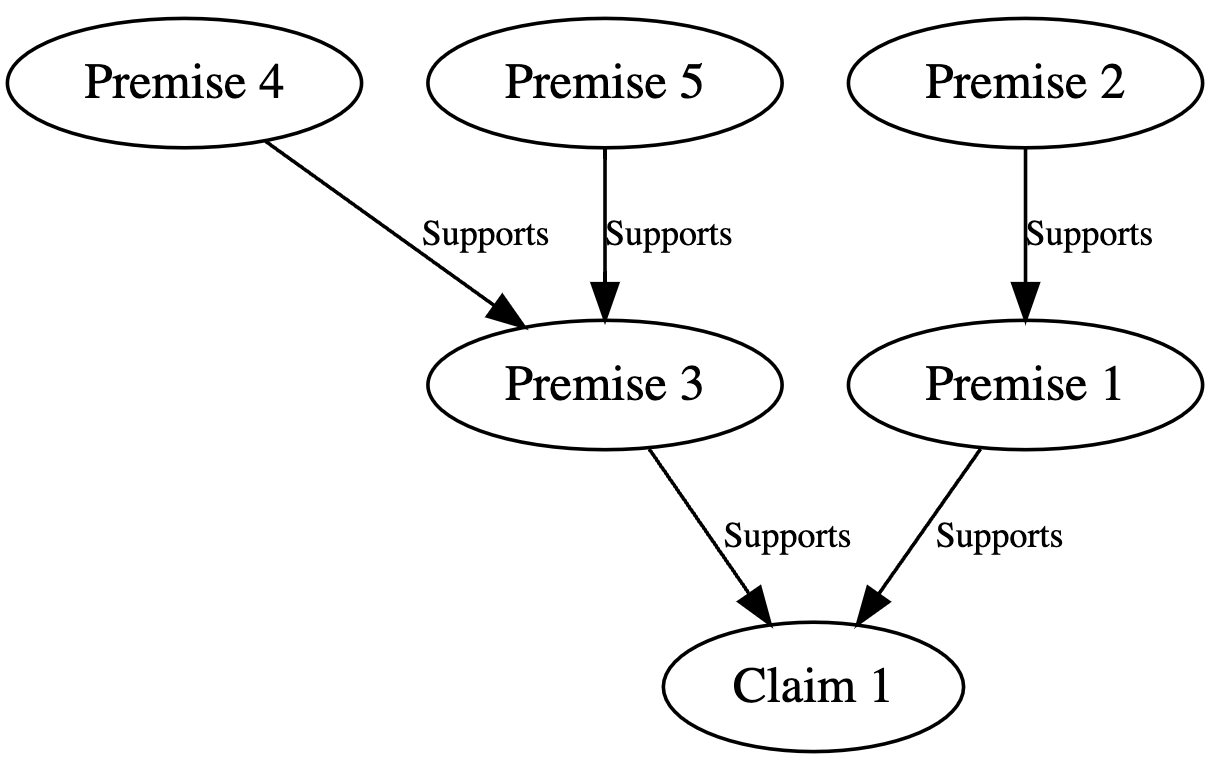}
    \caption{Relations between the argumentative components of the example introduced in \S \ref{sec:appendix_argument_structure}}.
    \label{fig:argument_relations}
\end{figure}

\begin{figure}
    \centering
    \includegraphics[scale=0.25]{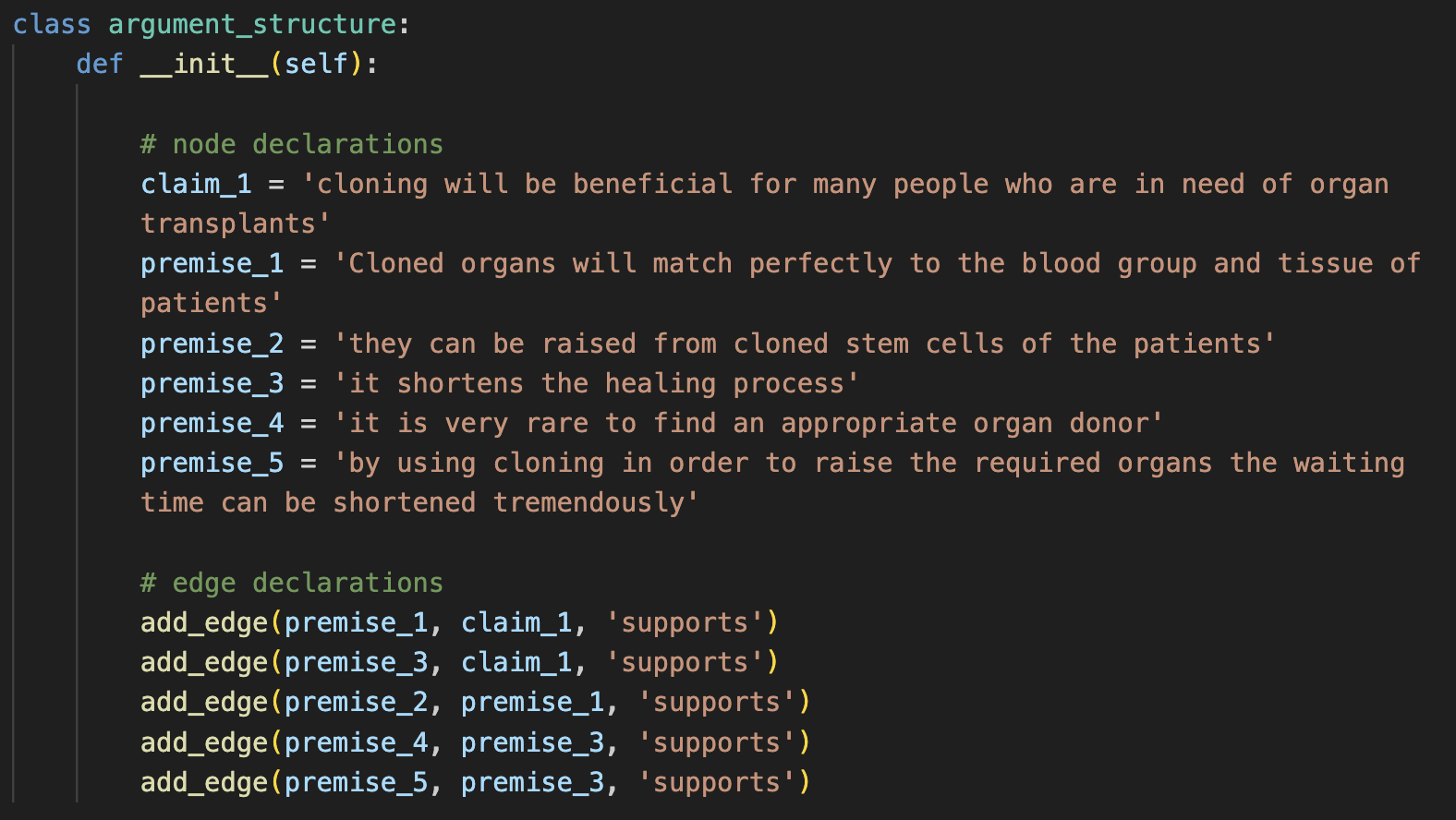}
    \caption{Programming script prompt used for argument structure extraction}
    \label{fig:argument_prompt}
\end{figure}

The example in Table \ref{tab:argument_text_annotated} shows the different argumentative components in a text along with their categories. The objective of argument structure extraction entails not only the identification of different argumentative components but also the prediction of support or attack relations between them. The argumentative relations between the components is shown in Figure \ref{fig:argument_relations}. The equivalent programming script representation of the aforementioned structure is shown in Figure \ref{fig:argument_prompt}.

Now, we provide some information on the datasets used to evaluate various approaches. We considered the following 3 datasets. \textbf{(1) \textsc{Essays}}~\cite{stab2017parsing} that consists of essays obtained from \url{essaysforum.com}. Each argumentative component within the essays is labeled at a sub-sentence level as either a premise, claim, or major claim. The relationships between these components are labeled as either attack or support. Test split consisting of $80$ datapoints is used for evaluation. We randomly select $7$ data points from the training split in the few-shot prompt. 
\textbf{(2) \textsc{AbstRCT}}~\cite{mayer2020transformer} is constructed by annotating the argumentative structure in the abstracts of PubMed 
articles on Randomized Controlled Trial of diseases. For our few-shot set, we randomly choose $11$ data points from the training split. The dataset includes three test splits: two from homogeneous data sources and one constructed by collecting data points from various sources, including the homogeneous ones. We focus on evaluating the performance of our model on the test split curated from various sources, which consists of $100$ data points.
\textbf{(3) \textsc{CDCP}}~\cite{park-cardie-2018-corpus} is obtained from a public forum where argumentative texts regarding proposed rules on Consumer Debt Collection Practices (CDCP) are annotated with argumentative components and the corresponding support relations. From the training set, we randomly select $7$ data points as our few-shot prompt. We then assess the performance of our model on the test split, which consists of $150$ data points.

\subsection{Task 2: Explanation Graph Generation: Additional information on prompt design and datasets}
\label{sec:appendix_explanation_graph_generation}
As we are focusing on the task of generating the commonsense structure, we assume that the stance is provided and prompt the model to generate the structure only as done in \citet{madaan-etal-2022-language}. We use the same prompt scheme as employed by \textsc{CoCoGen} in representing a graph as a programming script by directly adopting their implementation\footnote{\url{github.com/reasoning-machines/CoCoGen}}. 

\begin{figure}[h]
    \centering
    \includegraphics[scale=0.37]{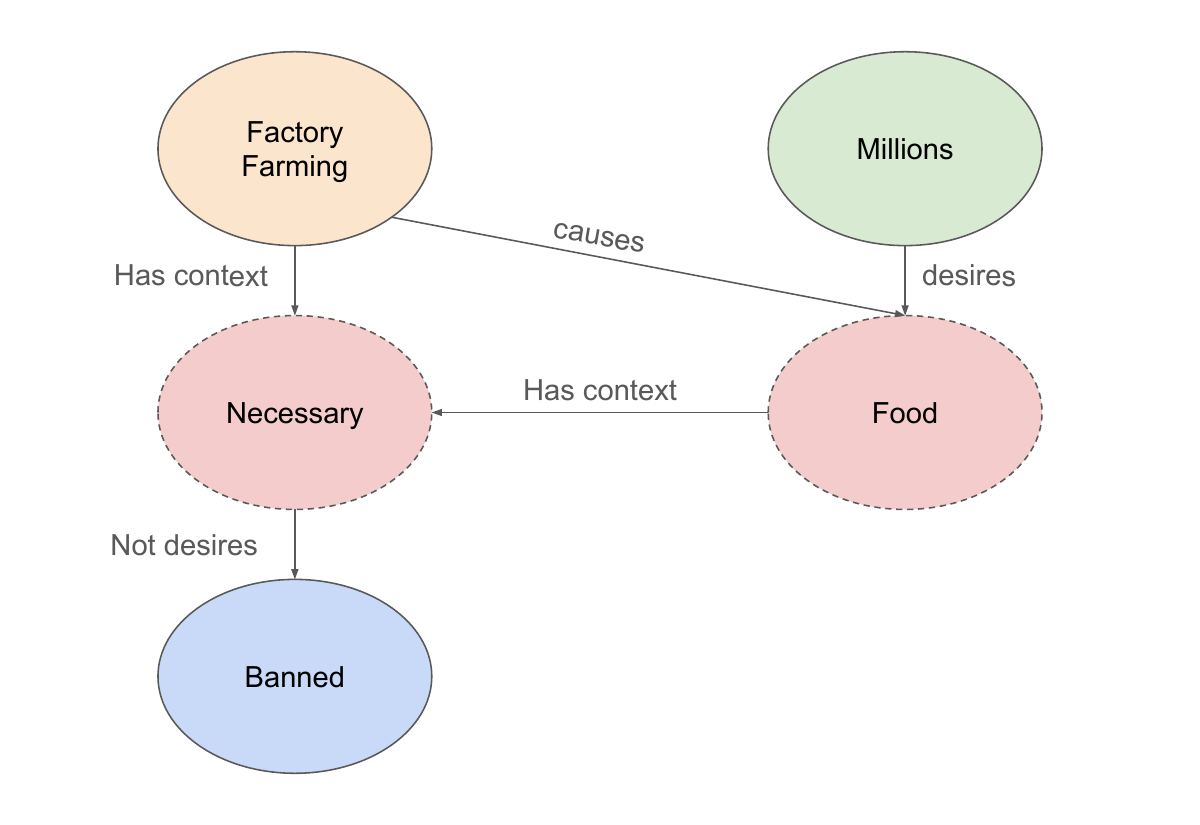}
    \caption{Explanation graph for the example shown in the Appendix \ref{sec:appendix_explanation_graph_generation}.}
    \label{fig:explagraph_example}
\end{figure}

In this implementation, $30$ few-shot instances were used to prompt the LLM and the approach was evaluated over the development split consisting of $396$ datapoints. An example explanation task for this task is shown in Figure \ref{fig:explagraph_example} for the following belief, argument and stance:
\begin{quote}
    \textbf{Belief:} Factory farming should not be banned.\\
    \textbf{Argument:} Factory farming feeds millions. \\
    \textbf{Stance:} Support
\end{quote}

\subsection{Task 3: Script Planning: Additional information on task, prompt design and datasets}
\label{sec:appendix_script_planning}

\begin{figure}
    \centering
    \includegraphics[scale=0.37]{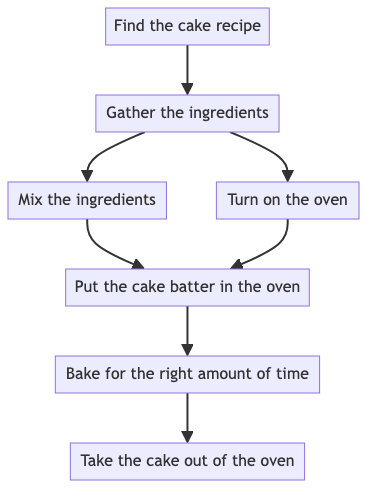}
    \caption{Script Planning for the goal \textit{"bake a cake"}. The steps shown in the figure are also provided as part of the input. The model has to predict directed relations between the steps that captures the temporal relations among them.}
    \label{fig:proscript_example}
\end{figure}

The input in \textsc{Proscript}~\cite{sakaguchi-etal-2021-proscript-partially} specifies the high-level goal to be achieved and the intermediate steps required to achieve the goal. The task involves inferring a sequence of dependency relationships among these steps, where each directed arrow indicates that the step at the arrow's head must be executed before the step at the tail. We utilize this dataset to evaluate the ability of various algorithms to automatically determine the order of operations needed to achieve the specified goal. We used $15$ few-shot instances for prompting and assessed the performance of various approaches over $100$ samples from the development dataset. An example of this datapoint is shown in Figure \ref{fig:proscript_example}.

\subsection{Task 4: Semantic Graph Generation: Additional information on task, prompt design and datasets}
\label{sec:additional_semantic_graph_construction}

\begin{table}[h]  
\centering  
\begin{tabular}{p{0.86\linewidth}} 
\midrule  
\textbf{Input Text:} While pop rock can trace its stylistic roots back to rock music, Reggae music evolved out of different musical genre, known as ska. Interestingly, the Train song, Mermaid, belongs to the genre of pop rock, but is also considered to be of the reggae genre as well  \\
\midrule
\textbf{Semantic Structure}: \textsc{("Mermaid Train song", "genre", "Pop rock"), ("Mermaid Train song", "genre", "Reggae"), ("Pop rock", "stylistic Origin", "Rock music"), ("Reggae", "stylistic Origin", "Ska")}\\
\bottomrule  
\end{tabular}
\caption{An example for Semantic Graph Generation.}
\label{tab:semantic_graph_generation}
\end{table} 

The goal of this task is to extract the semantic graph from an input graph, which is represented as a list of edges. Each edge in the graph consists of a subject, a property, and the type of property.~\cite{han2023pive}. An example of this task is shown in \ref{tab:semantic_graph_generation}.

To gauge the efficacy of our model for such a task, we consider following datasets: \textbf{(1) \textsc{Kelm}}~\cite{agarwal-etal-2021-knowledge}: This is a large scale synthetic dataset where each datapoint consists of a sentence in natural language and the corresponding semantic structure in the form of linearized Knowledge Graph (KG). Most of the graphs in this dataset contains at most $6$ edges. \textbf{(2) \textsc{WebNLG}}~\cite{gardent-etal-2017-webnlg}: The datapoints in this dataset were curated by sampling triples from the DBpedia~\cite{10.1007/978-3-540-76298-0_52}. The sentences describing their respective graphs were crafted using a wide range of lexicalization patterns. \textbf{(3) \textsc{GenWiki}}~\cite{jin-etal-2020-genwiki}: Unlike previous datasets, this one does not provide paired datapoints that map a natural language sentence to its corresponding semantic graph representation. However, a technique formulated by \citet{han2023pive} allows for the synthesis of pairwise annotated datasets, which we utilize in our assessments.

\section{Additional Analysis}
\subsection{Precision / Recall analysis for Argument structure extraction}
\label{sec:precision_recall_argument}
\begin{table*}[h]
    \centering
    \setlength\tabcolsep{2.5pt}
    \begin{tabular}{c|cccc|cccc|cccc}
        \midrule
        \multirow{2}{*}{Approach} & \multicolumn{4}{c|}{\textsc{Essays}} & \multicolumn{4}{c|}{\textsc{AbstRCT}} & \multicolumn{4}{c}{\textsc{CDCP}} \\
        & \textbf{C}(P) & \textbf{C}(R) & \textbf{R$_{50}$}(P) & \textbf{R$_{50}$}(R) & \textbf{C}(P) & \textbf{C}(R) & \textbf{R$_{50}$}(P) & \textbf{R$_{50}$}(R) & \textbf{C}(P) & \textbf{C}(R) & \textbf{R$_{50}$}(P) & \textbf{R$_{50}$}(R) \\
        \midrule
        \textsc{Greedy} & \bf 77.7 & 59.6 & 37.4 & 28.9 & \bf 86.9 & 81.9 & 50.1 & 61.4 & 55.0 & 52.6 & 13.4 & 21.0\\
         \model & 	74.0 & \bf 70.7 & \bf 40.5 & \bf 31.9 & 86.0 &	\bf 82.1 & \bf 53.7 & \bf 63.3 & \bf 55.9 & \bf 53.7 & \bf 14.4 & \bf 25.9\\
        \bottomrule
    \end{tabular}
    \caption{Component and Relation Identification precision and recall for \model and \textsc{Greedy}. P and R within the parentheses represent precision and recall respectively.}
    \label{tab:precision_recall_analysis}
\end{table*}

In order to empirically demonstrate the effectiveness of our algorithm in reducing errors, we compute the precision and recall in component and relation identification for argument structure extraction. This analysis not only allows us to assess the efficacy of our approach in filtering out false properties, but also in capturing genuine properties from multiple samples that would have otherwise been overlooked if only a single sample was relied upon. Instead of using the F$_1$-scores of the metrics \textbf{C} and \textbf{R$_{50}$} defined in \S\ref{sec:argument_structure_extraction}, we compute the precision and recall of these metrics under same definition. Specifically, the precision and recall along component identification is denoted by \textbf{C}(P) and \textbf{C}(R). A consistent notation is used for relation identification as well.

From the Table \ref{tab:precision_recall_analysis}, \model consistently improves the recall for component and relation identification across all datasets as it relies on multiple samples to formulate the final hypothesis, effectively addressing the issue of omitting true properties that would arise if relied on a single sample alone. Moreover, utilizing the consistencies among the samples leads to improved precision for relation identification, thereby helping reduce the number of spurious samples. While the precision for component identification is slightly impacted, adjusting the value of $\lambda_1$ allows us to achieve higher precision at the cost of slightly reduced recall.

\subsection{Impact of increasing the number of samples for other argument structure extraction tasks}
\label{sec:argument_extraction_samples}

\begin{figure}[H]
  \centering  
  \begin{subfigure}[b]{0.23\textwidth}  
    \includegraphics[width=\textwidth]{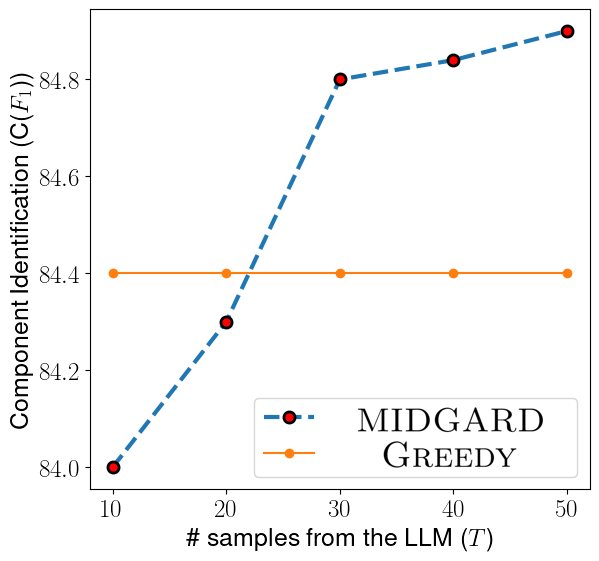}  
    \caption{Component Identification }  
    \label{fig:abstract_component_identification}  
  \end{subfigure}  
  \hfill  
  \begin{subfigure}[b]{0.23\textwidth}  
    \includegraphics[width=\textwidth]{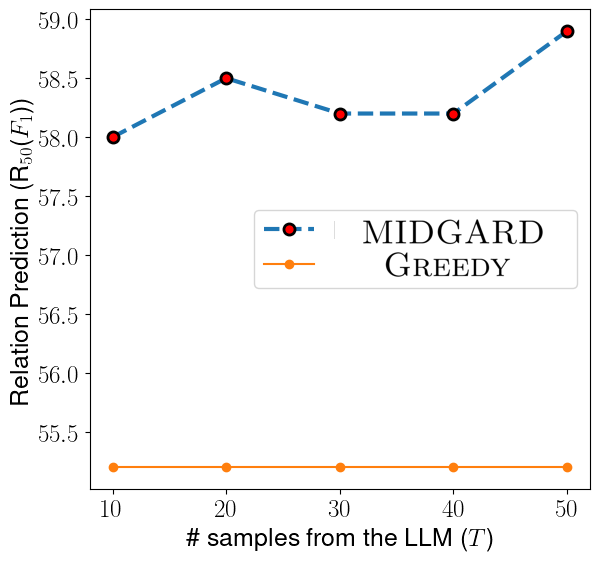}  
    \caption{Relation Prediction}  
    \label{fig:abstract_relation_prediction}  
  \end{subfigure}  
  \caption{Performance of \textsc{\model} in comparison with \textsc{Greedy} for the \textsc{AbstRCT} Dataset when the number of samples from the LLM is varied. Results averaged over $5$ different random seeds.}  
  \label{fig:abstract_performance}  
\end{figure} 

\begin{figure}[H]
  \centering  
  \begin{subfigure}[b]{0.23\textwidth}  
    \includegraphics[width=\textwidth]{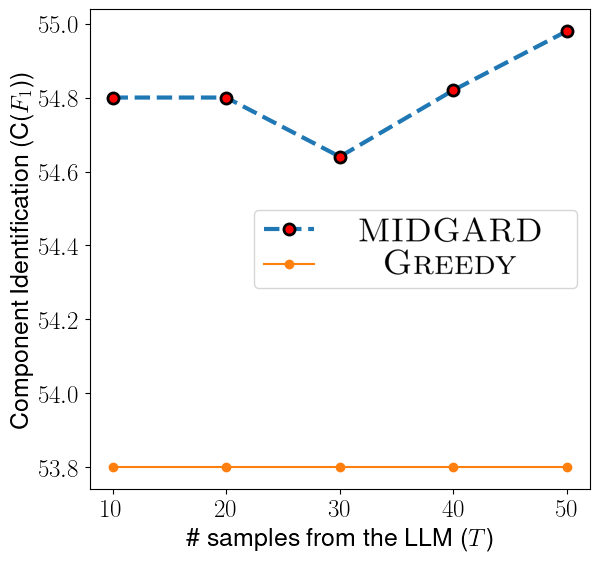}  
    \caption{Component Identification }  
    \label{fig:cdcp_component_identification}  
  \end{subfigure}  
  \hfill  
  \begin{subfigure}[b]{0.23\textwidth}  
    \includegraphics[width=\textwidth]{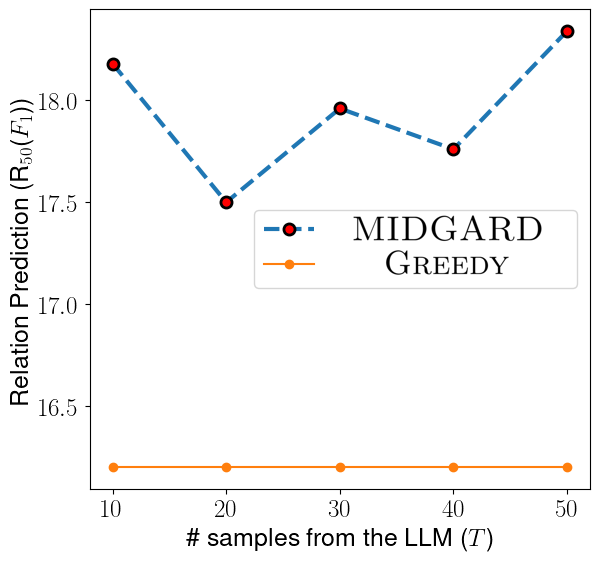}  
    \caption{Relation Prediction}  
    \label{fig:cdcp_relation_prediction}  
  \end{subfigure}  
  \caption{Performance of \textsc{\model} in comparison with \textsc{Greedy} for the \textsc{CDCP} Dataset when the number of samples from the LLM is varied. Results averaged over $5$ different random seeds.}  
  \label{fig:cdcp_performance}  
\end{figure}

\subsection{Impact of changing the number of few-shot examples for argument structure extraction}
\label{sec:argument_extraction_few_shot}
\begin{figure}[h]
  \centering  
  \begin{subfigure}[b]{0.23\textwidth}  
    \includegraphics[width=\textwidth]{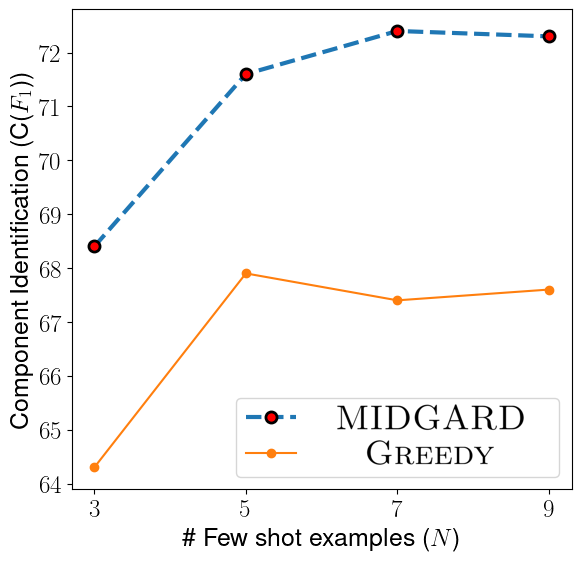}  
    \caption{Component Identification }  
    \label{fig:essay_few_shot_component_identification}  
  \end{subfigure}  
  \hfill  
  \begin{subfigure}[b]{0.23\textwidth}  
    \includegraphics[width=\textwidth]{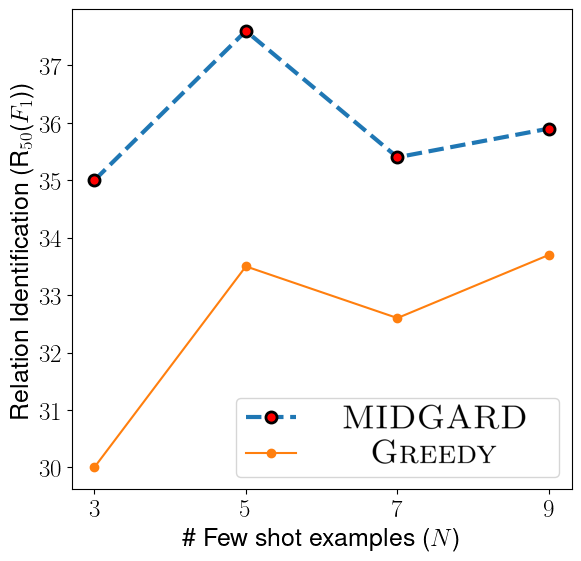}  
    \caption{Relation Prediction}  
    \label{fig:essay_few_shot_relation_identification}  
  \end{subfigure}  
  \caption{Performance of \textsc{\model} in comparison with \textsc{Greedy} for the \textsc{Essays} Dataset when the number of few shot examples ($N$) is varied. Results averaged over $5$ different random seeds.}  
  \label{fig:essay_performance_few_shot}  
\end{figure} 

\begin{figure}[H]
  \centering  
  \begin{subfigure}[b]{0.23\textwidth}  
    \includegraphics[width=\textwidth]{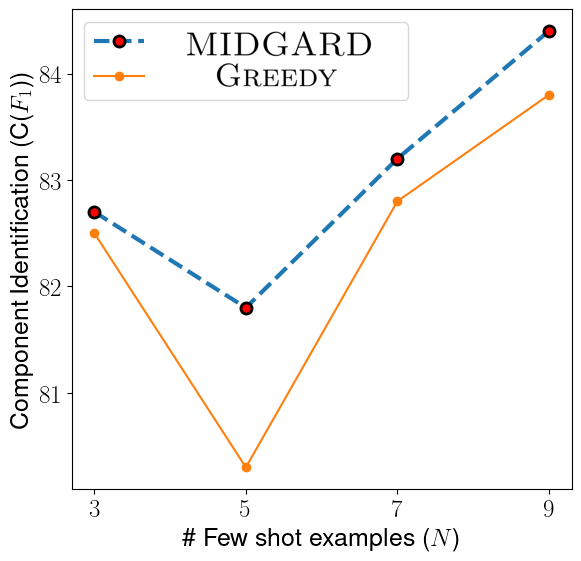}  
    \caption{Component Identification }  
    \label{fig:abstract_few_shot_component_identification}  
  \end{subfigure}  
  \hfill  
  \begin{subfigure}[b]{0.23\textwidth}  
    \includegraphics[width=\textwidth]{plots/abstract_few_shot_component_identification.png}  
    \caption{Relation Prediction}  
    \label{fig:abstract_few_shot_relation_identification}  
  \end{subfigure}  
  \caption{Performance of \textsc{\model} in comparison with \textsc{Greedy} for the \textsc{AbstRCT} Dataset when the number of few shot examples ($N$) is varied. Results averaged over $5$ different random seeds.}  
  \label{fig:abstract_performance_few_shot}  
\end{figure}

\begin{figure}[H]
  \centering  
  \begin{subfigure}[b]{0.23\textwidth}  
    \includegraphics[width=\textwidth]{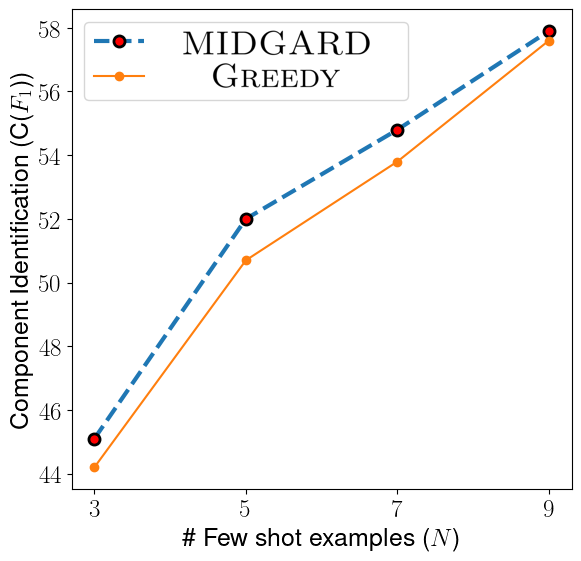}  
    \caption{Component Identification }  
    \label{fig:cdcp_few_shot_component_identification}  
  \end{subfigure}  
  \hfill  
  \begin{subfigure}[b]{0.23\textwidth}  
    \includegraphics[width=\textwidth]{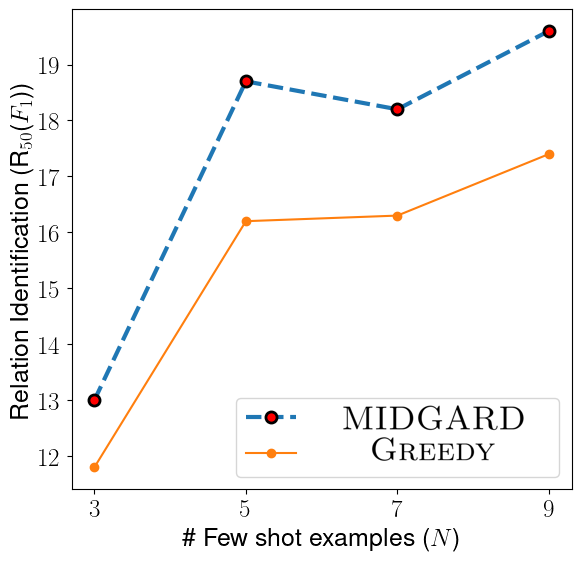}  
    \caption{Relation Prediction}  
    \label{fig:cdcp_few_shot_relation_identification}  
  \end{subfigure}  
  \caption{Performance of \textsc{\model} in comparison with \textsc{Greedy} for the \textsc{CDCP} Dataset when the number of few shot examples ($N$) is varied. Results averaged over $5$ different random seeds.}  
  \label{fig:cdcp_performance_few_shot}  
\end{figure}

We assess the effectiveness of our approach for different numbers of few-shot instances ($N \in \{3, 5, 7, 9\}$) in the context of argument structure extraction when $10$ samples are used from the LLM. As shown in Figure~\ref{fig:essay_performance_few_shot}, \model consistently enhances the performance of \textsc{Greedy} approach across different numbers of few shot examples. The plots for \textsc{AbstRCT} and \textsc{CDCP} are shown in Figure~\ref{fig:abstract_performance_few_shot} and Figure~\ref{fig:cdcp_performance_few_shot} respectively.

\subsection{Comparison with Nucleus Sampling}
\label{sec:appendix_nucleus_sampling}
While our evaluations considered \textsc{Greedy} decoding, we also compare against the \textsc{Nucleus} decoding~\cite{Holtzman2020The}, a popular technique to combat neural text degeneration, for the task of argument structure extraction in \textsc{Essays}. As shown in the Table \ref{tab:nucleus_sampling}, the application of \textsc{Nucleus} decoding degrades the performance significantly for both the considered LLMs.

\begin{table}[h]
    \centering
    \begin{tabular}{cccc}
        \toprule
        \textbf{Approach} & C & R$_{100}$ & R$_{50}$ \\
        \midrule
        \multicolumn{4}{c}{LLM: \texttt{gpt-35-turbo}} \\
        \midrule
        \textsc{Greedy} & 67.4 & 21.5 & 32.6 \\
        \textsc{Nucleus} & 64.1 & 19.2 & 31.2 \\
        \model & \bf 72.3 & \bf 23.5 & \bf 35.4 \\
        \midrule
        \multicolumn{4}{c}{LLM: \textsc{Code-Llama}} \\
        \midrule
        \textsc{Greedy} & 56.3 & 9.3 & 21.4 \\
        \textsc{Nucleus} & 48.0 & 6.7 & 16.7 \\
        \model & \bf 60.3 & \bf 11.0 & \bf 24.5 \\
        \bottomrule
    \end{tabular}
    \caption{Comparison of different approaches on gpt-35-turbo and Code-LLAMA models.}
    \label{tab:nucleus_sampling}
\end{table}

\subsection{Performance for \texttt{gpt-4}}
\label{sec:appendix_gpt4_evaluation}
Due to the prohibitive expense associated with \texttt{gpt-4}, we were limited in assessing its performance across all tasks. However, we have successfully evaluated its capabilities on the Argument Structure Extraction task using a selective subset of $20$ data points from the Essays Dataset. This specific evaluation thoroughly addresses both the identification of components (node evaluation) and the prediction of relations (edge evaluation), offering a more comprehensive analysis compared to other tasks. For the Essays dataset, which includes $80$ test data points, the estimated cost for GPT-4 analysis across 5 random instances of few-shot training examples could exceed $\$1000$. Given that other argument structure extraction datasets comprise over 80 test points, the expected inference costs would significantly increase. The results for the \textsc{Essays} dataset are tabulated in Table \ref{tab:gpt4_evaluation}.

\begin{table}[h]
    \centering
    \begin{tabular}{cccc}
        \toprule
        \textbf{Approach} & C & R$_{100}$ & R$_{50}$ \\
        \midrule
        \multicolumn{4}{c}{LLM: \texttt{gpt-4}} \\
        \midrule
        \textsc{Greedy} & 77.1 & 30.7 & 40.9 \\
        \model & \bf 78.1 & \bf 32.9 & \bf 42.8 \\
        \bottomrule
    \end{tabular}
    \caption{Comparison of different approaches implemented on \texttt{gpt-4} for \textsc{Essays} Dataset}
    \label{tab:gpt4_evaluation}
\end{table}

\subsection{Hyperparameters}
\label{sec:hyperparameter_analysis}
In this experiment, we vary $\lambda_1, \lambda_2 \in \{0.0, 0.1, 0.2, \dots , 1.0\}$ and compute the performance of component identification and relation prediction on \textsc{Essays}, and compare with that of hyperparameters automatically estimated (see Appendix \ref{sec:hyperparameter_selection} for more details). Specifically, when varying $\lambda_1$, we set $\lambda_2$ to 1 in order to include all nodes in the hypothesis and focus solely on studying the influence of $\lambda_1$ on relation prediction. An analogous step is repeated to study the influence of $\lambda_2$ on component identification. 

\begin{figure}
  \centering  
  \begin{subfigure}[b]{0.23\textwidth}  
    \includegraphics[width=\textwidth]{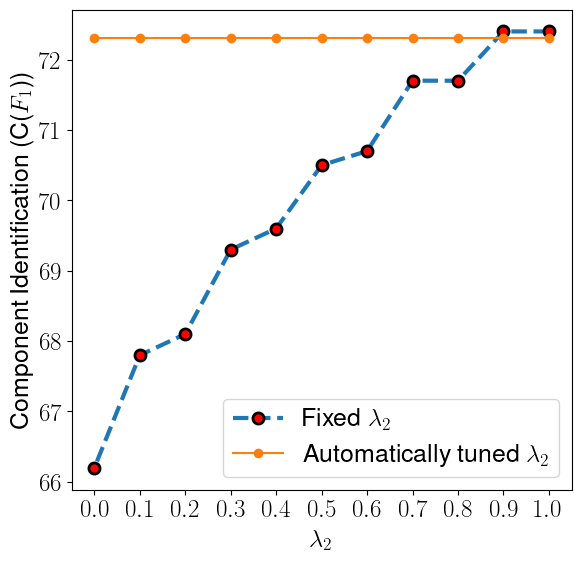}  
    \caption{Component Identification }  
    \label{fig:essay_threshold_analysis_node}  
  \end{subfigure}  
  \hfill  
  \begin{subfigure}[b]{0.23\textwidth}  
    \includegraphics[width=\textwidth]{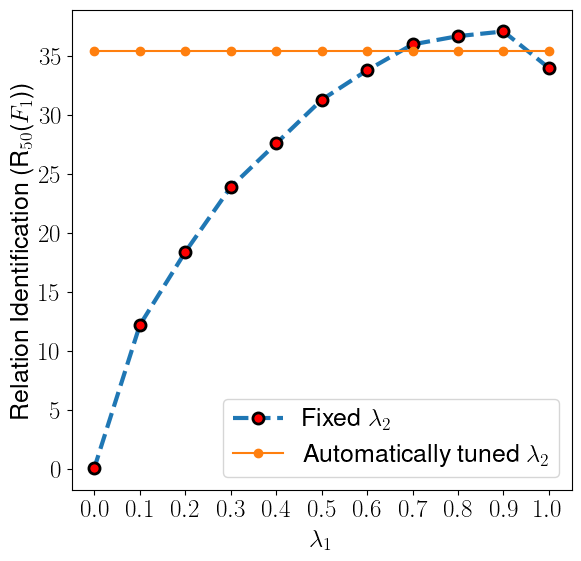}  
    \caption{Relation Prediction}  
    \label{fig:essay_threshold_analysis_edge}  
  \end{subfigure}  
  \caption{Assessing the performance of our algorithm for different values of $\{\lambda_1, \lambda_2\}$ and comparing it with that of automatically estimated hyperparameters. Results averaged over $5$ different random seeds.}  
  \label{fig:essay_threshold_analysis}  
\end{figure} 

In Figure~\ref{fig:essay_threshold_analysis}, we observe that the automatically estimated hyperparameter ($\lambda_2$) for component identification is near optimal performance. However, there is room for improvement in selecting $\lambda_1$.
Additionally, we find that the optimal values for both hyperparameters are above $0.5$, suggesting that the description length of insertion is greater than that of deletion, as discussed in Section \ref{sec:eqn_derivation}.

\section{Intuitive explanation for having unequal description lengths with addition versus deletion}
 
To define a single deletion, it requires $\propto \log_2(|E(\mathcal{G})|)$ to specify the edge to be deleted from $\mathcal{G}$. On the other hand, to describe the edge to be added one needs to spend $\propto \log_2(|\mathbf{E}|)$ bits. Clearly, $\log_2(|\mathbf{E}|) \geq \log_2(|E(\mathcal{G})|)$ as $\mathbf{E} \supseteq E(\mathcal{G})$.

\end{document}